\documentclass{article}

\usepackage{liquid_style}
\usepackage{liquid_color}
\usepackage{booktabs}
\usepackage{array}
\usepackage{tabularx}
\usepackage{pgfplots}
\pgfplotsset{compat=1.18}
\usetikzlibrary{arrows.meta, positioning, shapes.geometric}

\newcommand{\RepoRowURL}[2]{%
  \raisebox{-0.15\height}{\includegraphics[height=1.30em]{#1}}%
  & \href{#2}{\texttt{#2}}\\[-0.05em]  
}

\title{LFM2 Technical Report}

\author{
  Liquid AI Team\footnote{Please cite the author as ``Liquid AI (2025)''. See Section~\ref{sec:contributors} (Authors) for the list of contributors.}
}
\date{} 

\usepackage{hyperref}

\newcommand{\reportdate}{\today} 

\newcommand{\correspondingemail}{mathias@liquid.ai}

\let\origfootrule\footrule
\renewcommand{\footrule}{\iffootnote{}{\origfootrule}}

\begin{document}
\maketitle

\begin{center}
\setlength{\tabcolsep}{4pt} 
\renewcommand{\arraystretch}{1.05} 
\begin{tabular}{@{}>{\centering\arraybackslash}m{1.6em}@{\hspace{0.35em}}l@{}}
\RepoRowURL{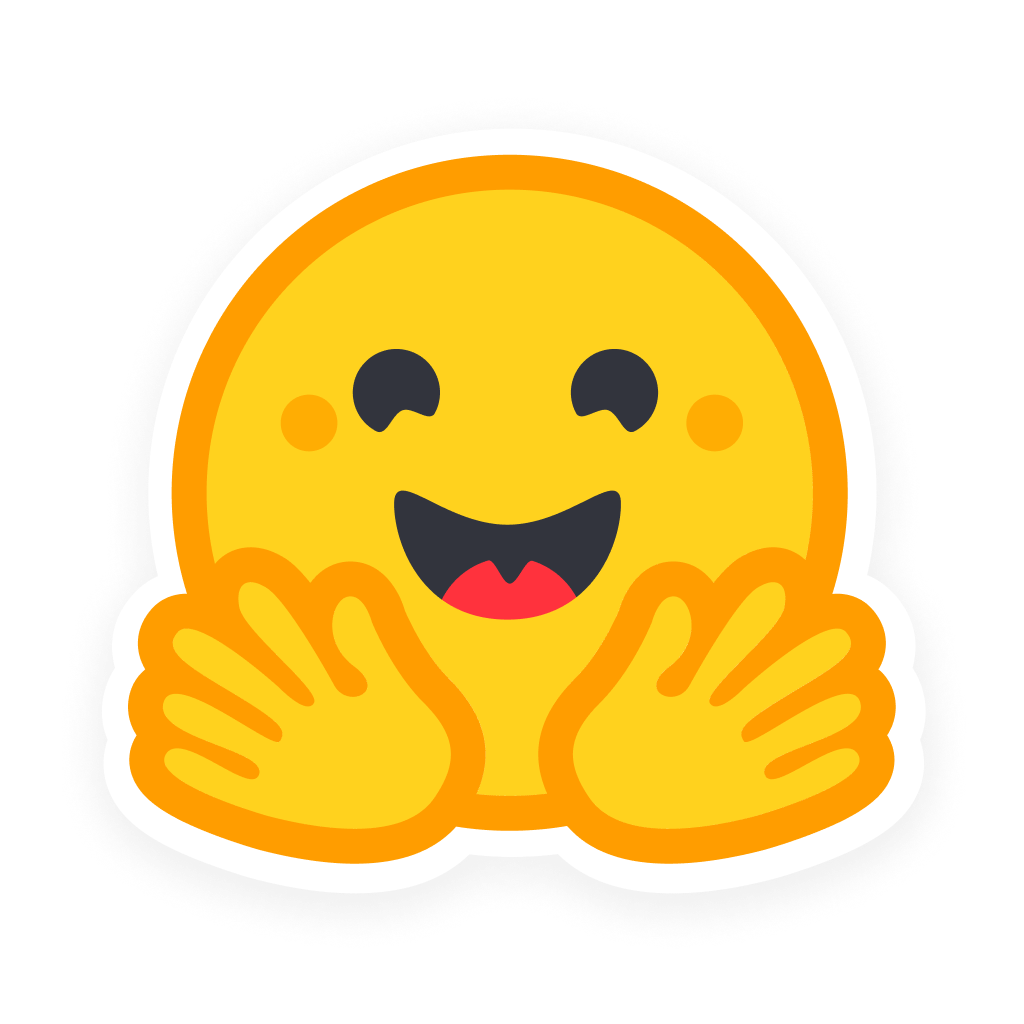}{https://huggingface.co/LiquidAI}
\end{tabular}
\end{center}
\vspace{1.2\baselineskip} 

\begin{abstractbox}
We present LFM2, a family of Liquid Foundation Models designed for efficient on-device deployment and strong task capabilities. Using hardware-in-the-loop architecture search under edge latency and memory constraints, we obtain a compact hybrid backbone that combines gated short convolutions with a small number of grouped query attention blocks, delivering up to 2$\times$ faster prefill and decode on CPUs compared to similarly sized models. The LFM2 family covers 350M–8.3B parameters, including dense models (350M, 700M, 1.2B, 2.6B) and a mixture-of-experts variant (8.3B total, 1.5B active), all with 32K context length. LFM2’s training pipeline includes a tempered, decoupled Top-K knowledge distillation objective that avoids support mismatch; curriculum learning with difficulty-ordered data; and a three-stage post-training recipe of supervised fine-tuning, length-normalized preference optimization, and model merging. Pre-trained on 10–12T tokens, LFM2 models achieve strong results across diverse benchmarks; for example, LFM2-2.6B reaches 79.56\% on IFEval and 82.41\% on GSM8K. We further build multimodal and retrieval variants: LFM2-VL for vision–language tasks, LFM2-Audio for speech, and LFM2-ColBERT for retrieval. LFM2-VL supports tunable accuracy–latency tradeoffs via token-efficient visual processing, while LFM2-Audio separates audio input and output pathways to enable real-time speech-to-speech interaction competitive with models 3$\times$ larger. LFM2-ColBERT provides a low-latency encoder for queries and documents, enabling high-performance retrieval across multiple languages. All models are released with open weights and deployment packages for ExecuTorch, llama.cpp, and vLLM, making LFM2 a practical base for edge applications that need fast, memory-efficient inference and strong task capabilities.

\vspace{0.7\baselineskip} 
{\footnotesize
\setlength{\tabcolsep}{3pt}%
\begin{tabular}{@{}l l@{}}
\textbf{Publication Date:} & \reportdate \\
\textbf{Correspondence:}   & \href{mailto:\correspondingemail}{\texttt{\correspondingemail}}
\end{tabular}
}
\end{abstractbox}

\vspace{1.2\baselineskip} 

\begin{figure}[htbp]
    \centering
    \includegraphics[width=1\textwidth]{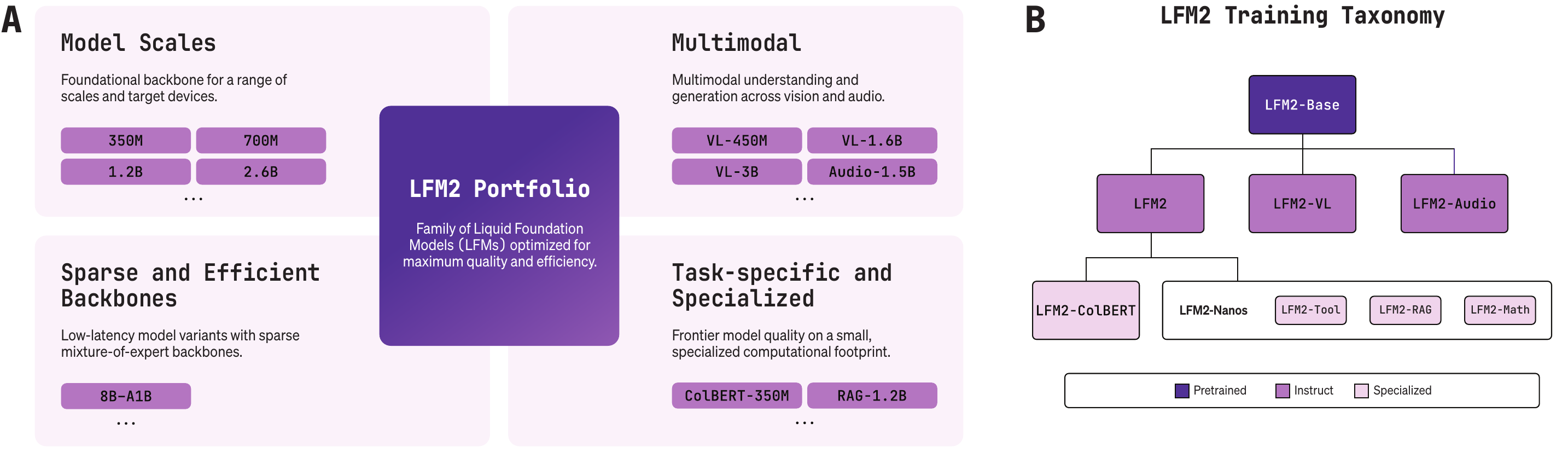}
    \caption{\textbf{The LFM2 Portfolio}. (A) We present a family of Liquid Foundation Models (LFMs) across a suite of scales, modalities, and edge capabilities. (B) The taxonomy of the LFM2 portfolio steps from co-designed architectures and pre-trained base models, to multi-modality, and specialized downstream tasks.}
    \label{fig:lfm2_overview}
\end{figure}

\section{Introduction}
\label{sec:intro}
Generative models that can be natively deployed on-device have the potential to bring transformative impact across sectors and industries \citep{xu2024device}. Applications such as voice assistants, local copilots, and agentic workflows that plan, call tools, and read sensors in tight loops must run under strict latency, memory, and energy budgets on phones, tablets, laptops, and system-on-chip (SoC) platforms. In these settings, time-to-first-token (TTFT), stable inter-token latency, and private or offline execution are not optional features, but rather hard constraints. Although recent open model families have improved efficiency at larger scales, targeting tens of billions of parameters and accelerator-heavy deployments~\citep{liu2024deepseek, yang2025qwen3,team2025gemma}, significant needs remain in the small-model, edge-first regime. We seek models that lead in quality, speed, and memory efficiency on CPUs and heterogeneous NPUs, while remaining practical to pre-train, post-train, and deploy widely.

Here, we introduce \textbf{LFM2}, the second generation of Liquid Foundation Models (LFMs) that are optimized explicitly for on-device deployment (Figure \ref{fig:lfm2_overview}). The LFM2 design is \emph{edge-first}: we co-design architecture, pre-training, and post-training around the objective of maximizing downstream quality subject to device-side latency and peak memory constraints. We release dense model checkpoints (\mbox{350M–2.6B}) with 32K context, an 8.3B mixture-of-experts (MoE) model with 1.5B active parameters, multimodal vision-language and audio-language variants, as well as a late-interaction retrieval model. We also release the initial series of LFM2-Nanos: LFM2 models that were further pre-trained and post-trained for domain or task-specific use cases such as data extraction, tool/function calling, retrieval augmented generation (RAG), and mathematical reasoning. All LFM2 models ship with open weights and deployment guides for Transformers \citep{wolf2020transformers}, \texttt{llama.cpp}, ExecuTorch, and vLLM \citep{kwon2023efficient}, along with quantized variants suited to edge runtimes\footnote{See the LiquidAI organization on Hugging Face: \url{https://huggingface.co/LiquidAI}.}. 
Key aspects of the LFM2 family include:
%
\begin{itemize}
  \item \textbf{Edge-first backbone for small, fast models.} A hardware-in-the-loop architecture search selects a minimal hybrid architecture that combines short-range, input-aware gated convolutions with grouped-query attention (GQA) in a layout tuned for quality under strict speed and memory budgets.
  %
  Using this backbone, dense LFM2 models from 350M to 2.6B parameters (32K context) and an MoE variant (LFM2-8B-A1B) cover a range of quality-latency-memory targets for on-device applications. 
  \item \textbf{Pre-training and post-training for small model, on-device workflows.} LFM2 checkpoints undergo a 10-12T token pre-training phase and a long-context mid-training phase. To improve small-model quality without prohibitive cost, we introduce a tempered, decoupled Top-K distillation objective that avoids support mismatch. A three-stage post-training pipeline improves performance and robustness at small model scales.

  \item \textbf{Native multimodality.} The LFM2 vision-language model, \textbf{LFM2-VL}, augments the language backbone with a SigLIP2~ \citep{tschannen2025siglip} vision encoder and a lightweight connector, designed to enable flexible accuracy-latency trade-offs on device. The audio-language model, \textbf{LFM2-Audio}, turns the backbone into an end-to-end audio–text model that ingests continuous audio features and autoregressively generates either text or discrete audio tokens, enabling low-latency speech assistants and transcription/translation in a single model. 
  \item \textbf{Multi- and cross-lingual information retrieval.} \textbf{LFM2-ColBERT} adds a dense module on top of the backbone, yielding a low-latency encoder for both queries and documents. It follows a late-interaction paradigm that uses a max-similarity operator \citep{khattab2020colbert}, enabling high-performance retrieval across multiple languages.
  \item \textbf{Strong performance with top efficiency}. Across sizes and applications, LFM2 models achieve strong tradeoffs between quality, latency, and memory, all critical considerations for edge deployment. On CPUs, LFM2 dense checkpoints deliver up to $\sim$2$\times$ prefill and decode speedups versus similarly sized baselines while maintaining or improving accuracy on benchmarks. LFM2-8B-A1B attains 3–4B-class quality at \(\sim\)1.5B active parameters.

\end{itemize}

This report first details the architecture and hardware-in-the-loop search procedure (Section~\ref{sec:arch}) used to design the LFM2 backbone. Detailed descriptions of the pre-training pipeline, including the long-context mid-training and decoupled Top-K distillation objective, and the three-stage post-training pipeline are provided in Sections~\ref{sec:pretraining} and~\ref{sec:post_training}, respectively, followed by results on model performance. Moving from the LFM2 language backbone to specific applications, we introduce the LFM2-VL (Section~\ref{sec:vlm}) and LFM2-Audio (Section~\ref{sec:audio}) multimodal models, covering their design, training, and evaluation, as well as the LFM2-ColBERT-350M model for information retrieval. We conclude with an extended discussion of related work (Section \ref{sec:related_work}) and a discussion of LFM2's limitations, opportunities for development, and impact on edge deployment (Section~\ref{sec:conclusion}).

\section{Architecture}
\label{sec:arch}

\begin{figure}[htbp]
    \centering
    \includegraphics[width=0.8\textwidth,
        trim={0 4cm 0 5cm}, clip]{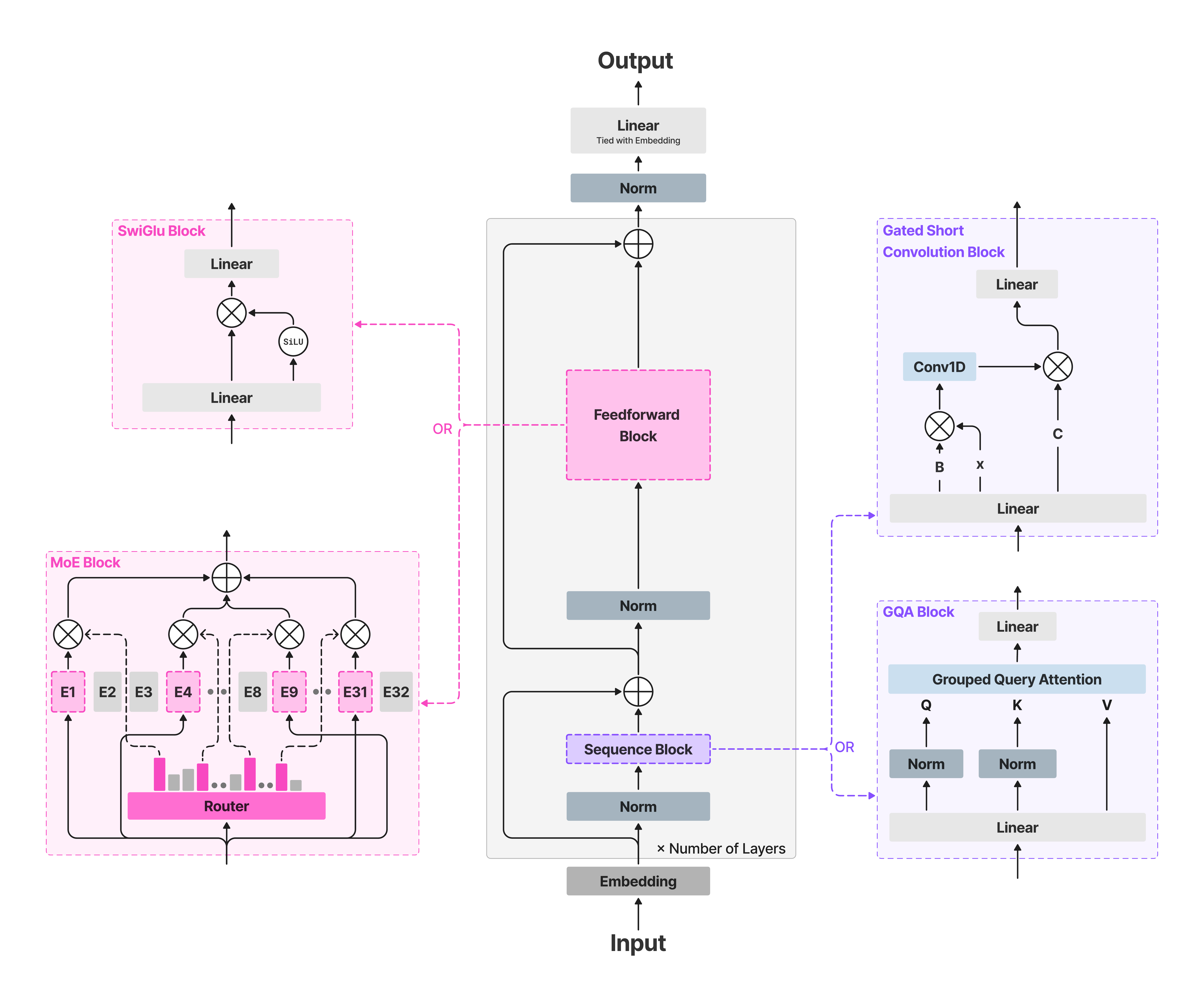}
    \caption{\textbf{LFM2 architecture.} The LFM2 architecture supports both dense and mixture-of-experts (MoE) variants. Note the MoE experts are also SwiGLU blocks.}
    \label{fig:lfm2_architecture}
\end{figure}

The LFM2 family instantiates the edge-first design introduced in Section \ref{sec:intro} with a minimal hybrid backbone designed via hardware-in-the-loop architecture search. We release 4 dense models: LFM2-350M, LFM2-700M, LFM2-1.2B, and LFM2-2.6B, as well as an MoE variant, LFM2-8B-A1B, which has 8.3B total parameters and 1.5B active parameters. Figure \ref{fig:lfm2_architecture} illustrates the shared backbone and the MoE extension. In this section, we describe the architecture search procedure (Section \ref{subsec:arch_opt}), the resulting dense backbone (Section \ref{subsec:dense}) and MoE (Section \ref{subsec:moe}) configurations, and their inference performance (Section \ref{subsec:inference}).

\subsection{Architecture Optimization Process}
\label{subsec:arch_opt}

The design objective for LFM2 is edge-first: maximize downstream quality while staying within tight device-side budgets on latency and peak memory on CPUs and heterogeneous NPUs. Recent efficient architectures~\citep{waleffe2024empirical, blakeman2025nemotron, team2025kimi} typically combine three ingredients: (i) alternative sequence operators such as linear attention variants or state space models (SSMs)~\citep{dao2024transformers, yang2025gated}, (ii) short-range convolutions (generally integrated in the alternative sequence block), and (iii) a non-trivial fraction of standard softmax attention layers to recover quality deficits (e.g., long-range retrieval abilities) observed in models without softmax attention layers~\citep{wen2025rnns, park2024can, waleffe2024empirical}. We explicitly search this design space with a hardware-in-the-loop search procedure and find that, under realistic edge latency and memory budgets, a minimal hybrid suffices: gated short convolutions for most layers plus a small minority of grouped-query attention (GQA) layers, without additional SSM or linear-attention operators.

\paragraph{Objectives and constraints.}
We frame architecture selection as a Pareto optimization over three axes:
\begin{enumerate}
    \item \textbf{Quality}: performance on an internal suite of $50+$ evaluations (spanning knowledge recall, multi-hop reasoning, instruction following, multilingual robustness, tool use, math, and long context performance) after training each candidate architecture on a reference dataset.
    \item \textbf{Latency}: time-to-first-token (TTFT) and p50/p95 decode latency (ms/token) at batch$=1$, as well as prefill throughput (tokens/s) on representative prompts.
    \item \textbf{Peak memory}: measured as maximum resident set size (RSS) during prefill and decode at target context windows (4K and 32K).
\end{enumerate}
Candidate architectures that violate device-side budgets on TTFT, decode latency, or peak memory are discarded.  The remaining candidates are ranked by hypervolume improvement on the quality–latency–memory Pareto frontier.

\paragraph{Search space.} We consider decoder-only stacks built from the following block families:
\begin{itemize}
  \item \textbf{Local context and subquadratic blocks}: gated short convolution blocks with varying kernel sizes, sliding-window attention~\citep{child2019generating}, and a family of sub-quadratic sequence blocks including linear attention variants~\citep{katharopoulos2020transformers,yang2024gated,yang2025gated}; 
  state-space variants such as S4~\citep{gu2022efficiently}, Liquid-S4~\citep{hasani2023liquid}, S5~\citep{smith2023simplified}, RTF~\citep{parnichkun2024state}, Mamba~\citep{gu2022efficiently}, and Mamba2~\citep{dao2024transformers}; Liquid-Time Constant networks such as CfC~\citep{hasani2022closed}, as well as internal variants of efficient sequence blocks. These blocks typically combine a depthwise short convolution with a longer-range linear attention/SSM component~\citep{fu2023hungry,poli2023hyena,gu2024mamba}. The search space includes variants that keep only the short convolution submodule (i.e., the gated short convolution block) as well as variants that retain the full hybrid operator. This allows the search to attribute performance gains to specific computational units within the overall operator. While linear attention and SSM variants can perform global processing of the input, we consider them to be in the same class as local context blocks due to their limitations in retrieval-intensive tasks~\citep{wen2025rnns, arora2024simple, blouir2024birdie, parnichkun2025quantifying} (see Section~\ref{subsec:rel_work_alt_seq} for a broader discussion).
  \item \textbf{Global context blocks}: grouped-query attention (GQA)~\citep{ainslie2023gqa} with varying group counts and head dimensions, augmented with QK-Norm~\citep{dehghani2023scaling}.
  \item \textbf{Position-wise blocks:} SwiGLU feed-forward blocks~\citep{shazeer2020glu} with expansion ratios chosen by search.
  \item \textbf{Layout}: interleaving patterns of local context blocks, global context blocks, position-wise blocks, and overall block counts under fixed parameter budgets, including options for shared weights and cache reuse.
  \item \textbf{MoE options}: per-layer sparse FFNs with varying width and expert granularity.
\end{itemize}

\paragraph{On-device profiling.}
Every candidate is compiled to the deployment stacks with identical settings (batch$=1$, fixed context windows at 4K/32K, and matched quantization/backends) and benchmarked on target devices:
\begin{itemize}
  \item \textbf{CPU path}: ExecuTorch (8da4w) and \texttt{llama.cpp} (Q4\_0) on Samsung Galaxy S24 Ultra (Qualcomm Snapdragon 8 Gen 3 SoC) and AMD Ryzen HX 370.
  \item \textbf{Accelerator path}: vLLM for single-request and online batching (used for sanity checks; the primary target remains on-device CPU deployment).
\end{itemize}
We record TTFT, prefill throughput (tokens/s), decode ms/token (p50/p95), and peak memory with identical prompts and tokenizer settings.

\paragraph{Evaluation and selection}
We consider quality estimation and deployability checks. Quality is measured via targeted training experiments scored on the internal evaluation suite. Deployability is assessed via the on-device profiling described above. Training experiments are run at scales large enough to surface meaningful quality differences, while device measurements are taken on representative edge hardware using the same runtimes at which we plan to ship. Configurations that preserve their advantage across both quality and deployability checks advance, while others are dropped early. Among feasible models, we retain those on the quality–latency–memory Pareto frontier and carry them forward in the search. 

Our earlier academic prototype (STAR)~\citep{thomasstar} explored a specific design space of operator/layout choices with an evolutionary search heuristic optimized on proxy signals (i.e., perplexity for quality, cache size for efficiency). In practice, these proxies do not transfer reliably to downstream task scores or device-level latency and memory, limiting their utility as optimization objectives.
By contrast, the LFM2 pipeline centers the \emph{objective}: downstream task scores and hardware-in-the-loop TTFT/latency/memory on release runtimes. In practice, we found this has a much larger impact than the particulars of the search space or choice of search heuristic.

\paragraph{Outcomes.}
Across size targets, the hardware-in-the-loop search repeatedly selects a \emph{minimal hybrid} architecture where most blocks are inexpensive gated short convolution blocks, interleaved with a small minority of GQA blocks. Under identical on-device performance budgets, augmenting these stacks (as in recent hybrid variants) with linear-attention, state-space, or additional convolution operators  does not improve aggregate quality on the evaluation suite and typically worsens device metrics. Empirically, the selected hybrids:
\begin{itemize}
  \item match or exceed the aggregate quality of attention-heavier and mixed (conv{+}linear/SSM/conv) baselines at the same budget;
  \item reduce decode latency (p50/p95) and increase prefill throughput at batch$=1$ under identical tokenizer, prompt, quantization, and backend settings;
  \item lower peak RSS at long context (4K/32K), consistent with reduced KV-cache versus attention-heavy layouts.
\end{itemize}

These results suggest that, in the on-device regime, most of the benefits attributed to recent hybrid SSM/linear-attention blocks can be captured by their short convolutional submodules plus a small number of global attention layers. We therefore carry forward designs that minimize global blocks while prioritizing inexpensive, gated short convolution blocks elsewhere. The released dense backbones (Section~\ref{sec:arch}) and the sparse-FFN placement in LFM2-8B-A1B (Section~\ref{sec:arch}) instantiate this recipe. Detailed speed and memory measurements appear in Section~\ref{subsec:inference}.

\subsection{LFM2 Dense Models}
\label{subsec:dense}

\paragraph{Block choices and layout.}
The LFM2 dense models instantiate the minimal hybrid repeatedly selected by the hardware-in-the-loop search (Section~\ref{subsec:arch_opt}): a majority of inexpensive, gated short convolution blocks interleaved with a minority of grouped-query attention (\textsc{GQA}) blocks, plus SwiGLU~\citep{shazeer2020glu} position-wise multi-layer perceptrons (MLPs). All layers use pre-norm RMSNorm~\citep{zhang2019root} and attention blocks use RoPE~\citep{su2024roformer} with QK-Norm~\citep{dehghani2023scaling}. 

\paragraph{Gated short convolution block.}
Given an input hidden sequence $\mathbf{h} \in \mathbb{R}^{L \times d}$ (batch dimension omitted for clarity), each gated short convolution block applies input-dependent multiplicative gating around a depthwise short convolution:
\[
(\mathbf{B},\,\mathbf{C},\,\tilde{\mathbf{h}})=\mathrm{Linear}(\mathbf{h}),\qquad
\mathbf{y}=\mathbf{B}\odot\tilde{\mathbf{h}},\qquad
\mathbf{z}=\mathrm{Conv}_k(\mathbf{y}),\qquad
\mathbf{o}=\mathrm{Linear}_{\text{out}}(\mathbf{C}\odot \mathbf{z}).
\]
Here $\mathrm{Linear}: \mathbb{R}^{d} \to \mathbb{R}^{3d}$ is a linear map applied position-wise across the sequence length, $L$, and whose output channels are split along the feature dimension into $(\mathbf{B},\mathbf{C},\tilde{\mathbf{h}})$ with $\mathbf{B},\mathbf{C},\tilde{\mathbf{h}} \in \mathbb{R}^{L \times d}$. The intermediate tensors $\mathbf{y},\mathbf{z},\mathbf{o}$ also lie in $\mathbb{R}^{L \times d}$. $\mathrm{Conv}_k: \mathbb{R}^{L \times d} \to \mathbb{R}^{L \times d}$ is a depthwise 1D convolution along the sequence with kernel size $k$, and $\odot$ denotes element-wise multiplication. $\mathrm{Linear}_{\text{out}}: \mathbb{R}^{d} \to \mathbb{R}^{d}$ is the linear output projection. This operator provides fast local mixing with excellent cache behavior on CPUs.

This operator is closely related to the short-range components that appear inside many recent efficient sequence blocks~\citep{fu2023hungry, poli2023hyena, gu2024mamba, dao2024transformers, yang2025gated}. The search results from Section \ref{subsec:arch_opt} can be viewed as an ablation of these hybrids in the on-device setting. Once a handful of GQA blocks are available to handle long-range retrieval, the inexpensive gated short convolution alone is sufficient to reach the best quality–latency–memory trade-off we observe, without additional linear attention/SSM/long convolution branches.

\paragraph{Attention and MLP.}
GQA reduces KV traffic by sharing keys/values across head groups while preserving multi-head queries~\citep{ainslie2023gqa}. Position-wise MLPs use SwiGLU with a size-dependent expansion ratio chosen by the search.

\paragraph{Tokenizer and special tokens.}
We use a byte-level BPE tokenizer~\citep{sennrich2016neural} with a 65{,}536-token vocabulary. We used the same pre-training dataset discussed in Section \ref{sec:pretraining} to train the LFM2 tokenizer, with a focus on encoding efficiency of the English, Japanese, Arabic, Korean, Spanish, French, and German languages. We additionally include JSON and other code-like data to improve tokenization of structured formats.

The tokenizer includes special tokens for fill-in-the-middle training objectives \citep{bavarian2022efficient}, tool calling, and the ChatML chat template. 

\paragraph{Released sizes.}
We release dense checkpoints at 350M, 700M, 1.2B, and 2.6B parameters, all with a $32{,}768$ token context window. Table~\ref{tab:lfm2_dense_specs} summarizes the main model hyperparameters.

\begin{table}[t]
\centering
\footnotesize
\setlength{\tabcolsep}{3pt}
\renewcommand{\arraystretch}{0.95}
\begin{tabular*}{\linewidth}{@{\extracolsep{\fill}} l r c c c c c c c c c @{}}
\toprule
& & \multicolumn{6}{c}{\textbf{Backbone}} & \multicolumn{3}{c}{\textbf{MoE}} \\
\cmidrule(lr){3-8}\cmidrule(lr){9-11}
\textbf{Model} & \multicolumn{1}{c}{Params T/A} &
Layers & $d_{\mathrm{model}}$ & FF dim & H/KV/$\mathrm{H}_\mathrm{size}$ & Attn. Blocks & Conv $k$ &
$E$ & Top-k & $\mathrm{FF}_{\mathrm{MoE}}$ \\
\midrule
LFM2-350M   & 350M/— & 16 & 1024 & 4608  & 16/8/64 & 6 & 3 & —  & — & — \\
LFM2-700M   & 700M/— & 16 & 1536 & 6912  & 24/8/64 & 6 & 3 & —  & — & — \\
LFM2-1.2B   & 1.2B/—  & 16 & 2048 & 8192  & 32/8/64 & 6 & 3 & —  & — & — \\
LFM2-2.6B   & 2.6B/—  & 30 & 2048 & 10752 & 32/8/64 & 8 & 3 & —  & — & — \\
\midrule
LFM2-8B-A1B & 8.3B/1.5B & 24 & 2048 & 7168 & 32/8/64 & 6 & 3 & 32 & 4 & 1792 \\
\bottomrule
\end{tabular*}
\caption{LFM2 model hyperparameters. “Backbone” columns list hyperparameters shared between dense and MoE models. T/A: total and active parameter counts. H/KV/$\mathrm{H}_\mathrm{size}$: Number of attention heads / KV groups / head size. “MoE” columns list experts per layer ($E$), active experts (Top-k), and per-expert FF size. For LFM2-8B-A1B, all layers except the first two are MoE.}
\label{tab:lfm2_dense_specs}
\end{table}

\subsection{LFM2 MoE}
\label{subsec:moe}

LFM2-8B-A1B (8.3B total, 1.5B active) targets the on-device regime where compute per token, not weight storage, dominates perceived latency. By activating only a sparse subset of experts per token, the model attains 3–4B-class quality (Table \ref{tab:large_models}) at roughly 1.5B-class decode cost. We keep the fast LFM2 hybrid backbone (Section~\ref{subsec:dense}) and replace dense MLPs with sparse MoE MLPs in most layers. For stability, the first two layers remain dense while all subsequent layers include an MoE block. Experts themselves are SwiGLU MLPs. Each MoE layer has 32 experts and selects the Top-k${=}4$ experts per token with a normalized sigmoid router and adaptive routing biases for load balancing~\citep{liu2024deepseek}. See
Table~\ref{tab:lfm2_dense_specs} for a summary of the main LFM2-8B-A1B settings.

\subsection{Inference Performance}
\label{subsec:inference}

\paragraph{Experimental setup.}
We evaluate the inference efficiency of both dense and MoE LFM2 models on two representative on-device CPU targets: a Snapdragon 8 Elite based Samsung Galaxy S25 smartphone (a newer model than we used for the architecture search in Section \ref{subsec:arch_opt}) and an AMD Ryzen AI~9 HX 370 laptop CPU. In both settings, we focus on latency-sensitive, single-stream usage with batch size equal to 1. All models are run with \texttt{llama.cpp}~\citep{gerganov_llama_cpp} using the Q4\_0 quantization format. For ease of comparison and reproducibility, we measure
\emph{prefill throughput} (prompt tokens processed per second) for 1K and 4K-token prompts and
\emph{decode throughput} (tokens generated per second) when producing 100 continuation tokens from 1K and 4K-token prefixes. Tables~\ref{tab:s25_inference} and~\ref{tab:hx370_inference} summarize the results for the LFM2 family alongside contemporary open baselines~\citep{yang2025qwen3, team2025gemma, grattafiori2024llama, bakouch2025smollm3, granite2025} at comparable parameter scales.

\begin{table}[H]
\centering
\small
\begin{tabular}{lrrrr}
\toprule
& \multicolumn{2}{c}{\textbf{Prefill throughput (tokens/s)}} & \multicolumn{2}{c}{\textbf{Decode throughput (tokens/s)}}\\
\cmidrule(lr){2-3}\cmidrule(lr){4-5}
\textbf{Model} & 1K & 4K & 1K & 4K \\
\midrule
LFM2-350M & 1,067 & 657 & 194.1 & 143.8\\
Granite-4.0-350M & 528 & 210 & 132.9 & 70.7\\
Granite-4.0-H-350M & 784 & 594 & 150.7 & 119.3\\
\midrule
LFM2-700M & 522 & 341 & 104.2 & 80.2\\
Qwen3-0.6B & 318 & 136 & 76.7 & 41.8\\
\midrule
LFM2-1.2B & 335 & 222 & 69.8 & 55.5\\
Llama-3.2-1B  & 229 & 130 & 54.6 & 37.8\\
Qwen3-1.7B & 140 & 98 & 39.7 & 26.9\\
Gemma-3-1B  & 377 & 295 & 67.5 & 67.1\\
Granite-4.0-1B & 147 & 79 & 41.3 & 25.0\\
Granite-4.0-H-1B & 186 & 159 & 46.1 & 44.1\\
\midrule
LFM2-2.6B & 143 & 116 & 33.8 & 30.0\\
LFM2-8B-A1B & 85 & 76 & 48.6 & 41.9\\
Llama-3.2-3B  & 79 & 51 & 24.2 & 15.8\\
Qwen3-4B  & 57 & 35 & 17.2 & 11.4\\
Gemma-3-4B  & 72 & 63 & 18.3 & 17.9\\
SmolLM3-3B & 98 & 66 & 26.1 & 19.2\\
Granite-4.0-Micro & 65 & 38 & 20.2 & 12.1\\
Granite-4.0-H-Micro & 83 & 77 & 24.9 & 23.5\\
Granite-4.0-H-Tiny & 86 & 86 & 39.2 & 37.4\\
\bottomrule
\end{tabular}
\caption{Prefill and decode throughput on the Samsung Galaxy S25 device with a Qualcomm Snapdragon 8 Elite SoC (batch size 1). Prefill columns report tokens per second when processing prompts of length 1K/4K tokens. Decode columns report tokens per second when generating 100 tokens starting from prefixes of length 1K/4K tokens. All models are run with llama.cpp and Q4\_0 format with CPU backend.}
\label{tab:s25_inference}
\end{table}

\paragraph{Smartphone-class CPU (Samsung Galaxy S25 with a Qualcomm Snapdragon 8 Elite SoC).}
On the S25 device (Table~\ref{tab:s25_inference}), LFM2 models are generally faster than similarly sized baselines across scales. 

In the 350M class, LFM2-350M yields about $2$--$3\times$ higher prefill throughput than Granite-4.0-350M and improves decode throughput by roughly $1.5$-$2.0\times$ across 1K and 4K contexts. Relative to the hybrid Granite-4.0-H-350M, LFM2-350M remains $\sim$10--35\% faster in both prefill and decode. At the 700M scale, LFM2-700M outperforms Qwen3-0.6B across all four settings. Prefill throughput improves by about $1.6$--$2.5\times$, while decode throughput improves by roughly $1.4$--$2\times$.

At the 1B scale, LFM2-1.2B improves over Granite-4.0-1B and Qwen3-1.7B by $2.3$--$2.8\times$ on prefill and $1.7$--$2.2\times$ on decode across both context lengths. Gemma-3-1B is the strongest 1B-class baseline in terms of raw throughput; however, LFM2-1.2B attains $0.75-0.9\times$ of its prefill throughput, while offering slightly higher 1K decode throughput and remaining within $20\%$ on 4K decode. In contrast, relative to Llama-3.2-1B, LFM2-1.2B is faster across all four measurements, with roughly $1.5$--$1.7\times$ higher prefill throughput and $1.3$--$1.5\times$ higher decode throughput.

In the 2--4B regime, the dense LFM2-2.6B model delivers about $2$--$3\times$ higher throughput than Qwen3-4B across prefill and decode, and also outperforms Gemma-3-4B, Granite Micro/H variants, SmolLM3-3B, and Llama-3.2-3B in both metrics. The MoE model LFM2-8B-A1B attains prefill throughput comparable to Granite-4.0-H-Tiny (another hybrid MoE), but with $1.1$--$1.3\times$ higher decode throughput, while providing roughly $2.8$--$3.7\times$ higher decode throughput than dense 4B-class baselines such as Qwen3-4B. We have found that there is still room for large improvements for MoEs on CPU, and we are developing improved kernels that better utilize CPU hardware.

\paragraph{Laptop-class CPU (AMD Ryzen HX 370).}
On the Ryzen AI~9 HX 370 CPU (Table~\ref{tab:hx370_inference}), absolute throughputs increase substantially compared to the S25, and the relative trends largely mirror the smartphone results.

\begin{table}[H]
\centering
\small
\begin{tabular}{lrrrr}
\toprule
& \multicolumn{2}{c}{\textbf{Prefill throughput (tokens/s)}} & \multicolumn{2}{c}{\textbf{Decode throughput (tokens/s)}}\\
\cmidrule(lr){2-3}\cmidrule(lr){4-5}
\textbf{Model} & 1K & 4K & 1K & 4K \\
\midrule
LFM2-350M & 7,534 & 5,540 & 207.7 & 170.5\\
Granite-4.0-350M & 5,499 & 2,938 & 225.0 & 166.2\\
Granite-4.0-H-350M & 4,568 & 4,250 & 134.7 & 123.2\\
\midrule
LFM2-700M & 4,214 & 3,311 & 134.6 & 118.4\\
Qwen3-0.6B & 3,594 & 2,204 & 116.0 & 53.4\\
\midrule
LFM2-1.2B & 2,784 & 2,302 & 99.7 & 89.0\\
Llama-3.2-1B  & 2,912 & 1,890 & 97.3 & 73.5\\
Qwen3-1.7B & 2,019 & 1,491 & 60.8 & 38.5\\
Gemma-3-1B  & 3,972 & 3,809 & 103.6 & 99.3\\
Granite-4.0-1B & 1,823 & 1,256 & 64.5 & 43.6\\
Granite-4.0-H-1B & 1,565 & 1,523 & 59.6 & 57.1\\
\midrule
LFM2-2.6B & 1,335 & 1,171 & 50.3 & 46.8\\
LFM2-8B-A1B & 1,320 & 1,185 & 74.9 & 69.1\\
Llama-3.2-3B  & 1,179 & 916 & 38.3 & 28.2\\
Qwen3-4B  & 861 & 628 & 31.0 & 22.2\\
Gemma-3-4B  & 1,119 & 1,054 & 33.0 & 31.6\\
SmolLM3-3B & 1,248 & 982 & 42.4 & 33.4\\
Granite-4.0-Micro & 929 & 608 & 36.7 & 28.9\\
Granite-4.0-H-Micro & 932 & 896 & 34.7 & 33.8\\
Granite-4.0-H-Tiny & 1,050 & 1,046 & 55.0 & 52.1\\
\bottomrule
\end{tabular}
\caption{Prefill and decode throughput on HX 370 CPU (batch size 1). Prefill columns report tokens per second when processing prompts of length 1K/4K tokens. Decode columns report tokens per second when generating 100 tokens starting from prefixes of length 1K/4K tokens. All models are run with llama.cpp and Q4\_0 format with CPU backend.}
\label{tab:hx370_inference}
\end{table}

\paragraph{Summary.}
Across both smartphone and laptop class CPUs, LFM2 models generally achieve substantial latency and throughput improvements over similarly sized open baselines. Together with the quality measurements presented in Section \ref{subsec:posttrain_eval}, these results support the edge-first design of the LFM2 backbone and its suitability for latency-sensitive on-device deployment (Figure \ref{fig:samsung-s25-pareto}).

\section{Pre-Training}
\label{sec:pretraining}

We pre-train LFM2 using a multi-stage pipeline designed for small, on-device models. During pre-training, we combine standard next-token prediction with a tempered, decoupled Top-K knowledge distillation objective (Section \ref{subsec:KD}), using an internal version of LFM1-7B as a teacher. This section details the pre-training data mixture (Section \ref{subsec:train_data}), the training stages (Section \ref{subsec:train_stages}), and the decoupled Top-K distillation objective (Section \ref{subsec:KD}).

\subsection{Training Data}
\label{subsec:train_data}
The LFM2 dense models are pre-trained on a mixture comprising roughly 75\% English text, 20\% multilingual text, and 5\% code. We prioritize Japanese, Arabic, Korean, Spanish, French, and German for multilingual coverage, with additional base support for Chinese, Italian, and Portuguese. The MoE model LFM2-8B-A1B uses a similar mixture, but with a heavier emphasis on code (60\% English, 25\% multilingual, 15\% code). For code data, 50\% of examples use a fill-in-the-middle (FIM) objective~\citep{bavarian2022efficient}.

\subsection{Training Stages}
\label{subsec:train_stages}
The released dense LFM2 model checkpoints are pre-trained for 10T tokens at a context length of 4,096 tokens. We then perform a mid-training phase on an additional 1T higher-quality tokens, including sources with naturally long context, using a 32,768-token context window and an accelerated learning-rate decay schedule. The released LFM2-8B-A1B checkpoint follows the same two-stage recipe but is trained for 12T tokens in the initial phase before the 1T-token long-context mid-training stage.

\subsection{Decoupled Top-K Knowledge Distillation}
\label{subsec:KD}

During pre-training, we leverage LFM1-7B as a teacher model in a knowledge distillation (KD) framework~\citep{hinton2015distilling}. To reduce storage and bandwidth, we distill the teacher distribution $P_T(x\mid x_c)$ for each token $x$ given its context $x_c$ to a student $P_S(x\mid x_c)$, using only the Top-K${=}32$ teacher logits per token. Naively applying the Kullback–Leibler (KL) divergence to a Top-K truncated teacher distribution, especially under temperature scaling, can cause support mismatch and unstable losses. We address this by decomposing the KL via the chain rule into (i) a binary term that matches the total probability mass assigned to the Top-K set, and (ii) a conditional KL within the Top-K, to which we apply temperature. This “decoupled, tempered Top-K” objective avoids support mismatch while preserving the benefits of temperature-scaled distillation.

Let \(\mathcal A\) be the vocabulary and \(\mathcal T(x_c)\subset\mathcal A\) be the teacher’s Top-K set for context \(x_c\). Define
\[
P_T(\mathcal T\mid x_c)=\sum_{x\in \mathcal T(x_c)}P_T(x\mid x_c),\quad
P_S(\mathcal T\mid x_c)=\sum_{x\in \mathcal T(x_c)}P_S(x\mid x_c),
\]
\[
P_T(x\mid \mathcal T,x_c)=\frac{P_T(x\mid x_c)}{P_T(\mathcal T\mid x_c)},\qquad
P_S(x\mid \mathcal T,x_c)=\frac{P_S(x\mid x_c)}{P_S(\mathcal T\mid x_c)}\quad (x\in\mathcal T).
\]

We denote applying a temperature \(\tau\) as
\[
p^{(\tau)}(x)=\frac{p(x)^{1/\tau}}{\sum_{y}p(y)^{1/\tau}},\qquad
D_{\mathrm{KL}}^{(\tau)}(p\|q)\;=\;\tau^2\,D_{\mathrm{KL}}\!\big(p^{(\tau)}\|q^{(\tau)}\big).
\]

The decoupled, tempered Top-K objective (per token) is
\begin{align}
\mathcal L_{\text{DTK}}(x_c)
&=
\underbrace{D_{\mathrm{KL}}\!\big(\mathrm{Bern}(P_T(\mathcal T\mid x_c))\,\big\|\,\mathrm{Bern}(P_S(\mathcal T\mid x_c))\big)}_{\mathcal L_B}
\\[-2pt]
&\quad+
\underbrace{P_T(\mathcal T\mid x_c)\,D_{\mathrm{KL}}^{(\tau)}\!\big(P_T(\cdot\mid \mathcal T,x_c)\,\big\|\,P_S(\cdot\mid \mathcal T,x_c)\big)}_{\mathcal L_T}
\end{align}
where \(\mathrm{Bern}(p)\) denotes the Bernoulli distribution with success probability \(p\). The first term $\mathcal{L}_B$ is a binary KL that trains the student to put the same \emph{total probability mass} into the teacher's Top-K set as the teacher does. The second term $\mathcal{L}_T$ is a conditional Top-K KL divergence that, given the next token lies in the Top-K, trains the student to match the teacher’s \emph{relative probabilities} among those \(K\) tokens. 

We apply temperature \textbf{only} to the conditional Top-K distributions inside $\mathcal L_T$. The Bernoulli membership term $\mathcal L_B$ and the Top-K mass $P_T(\mathcal T\mid x_c)$ remain untempered, which prevents the support mismatch that arises when tempering a Top-K truncated distribution over the full vocabulary. Note that with $\tau=1$, $\mathcal L_{\text{DTK}}$ is a lower bound to the full forward KL because it omits the teacher tail where logits are unavailable. For $\tau > 1$, this can be viewed as optimizing a tempered surrogate of this bound. Please see Appendix \ref{app:dtk} for the full derivation and more details. Finally, the overall $\mathcal L_{\text{DTK}}$ term is balanced with next-token cross-entropy on hard labels. See Section \ref{subsec:rel_work_distill} for an expanded discussion on related approaches such as \citet{zhao2022decoupled} and \citet{anshumann2025sparse}.

\section{Post-Training}
\label{sec:post_training}

\subsection{Overview}
\begin{figure}[htb]
    \centering
    \begin{tikzpicture}[
        node distance=1.3cm and 0.8cm,
        stage/.style={
            rectangle, rounded corners=3pt, 
            minimum width=1.8cm, minimum height=1.1cm,
            align=center, font=\footnotesize\bfseries,
            draw=liquidpurple,
            fill=white,
            line width=0.6pt
        },
        model/.style={
            ellipse, minimum width=1.4cm, minimum height=0.85cm,
            align=center, font=\scriptsize,
            draw=black,
            fill=white,
            line width=0.6pt
        },
        data/.style={
            rectangle, rounded corners=2pt,
            minimum width=1.2cm, minimum height=0.55cm,
            align=center, font=\scriptsize,
            draw=black,
            fill=white,
            line width=0.4pt
        },
        arrow/.style={-{Stealth[length=1.8mm]}, line width=0.5pt, draw=black},
        dashedarrow/.style={-{Stealth[length=1.8mm]}, line width=0.5pt, dashed, draw=black}
    ]

    \path (0,2) -- (0,0);

    \node[model] (base) {Base\\Model};
    \node[stage, right=of base, text=liquidpurple] (sft) {Stage 1\\SFT};
    \node[model, right=of sft] (sftout) {SFT\\Model};
    \node[stage, right=of sftout, text=liquidpurple] (pref) {Stage 2\\Alignment};
    \node[model, right=of pref] (prefout) {Aligned\\Models};
    \node[stage, right=of prefout, text=liquidpurple] (merge) {Stage 3\\Merging};
    \node[model, right=of merge] (final) {Final\\Model};

    \node[data, above=0.6cm of sft] (sftdata) {SFT Data};
    \node[data, above=0.6cm of pref, xshift=-0.7cm] (onpolicy) {On-policy};
    \node[data, above=0.6cm of pref, xshift=0.7cm] (offpolicy) {Off-policy};

    \node[font=\scriptsize\bfseries, text=black, above=0cm of onpolicy] (clair) {CLAIR};

    \draw[arrow] (base) -- (sft);
    \draw[arrow] (sft) -- (sftout);
    \draw[arrow] (sftout) -- (pref);
    \draw[arrow] (pref) -- (prefout);
    \draw[arrow] (prefout) -- (merge);
    \draw[arrow] (merge) -- (final);

    \draw[arrow] (sftdata) -- (sft);
    \draw[arrow] (offpolicy) -- (pref);
    \draw[arrow] (onpolicy) -- (pref);

    \draw[dashedarrow] (sftout.north) |- (onpolicy.west);

    \path (current bounding box.south) ++(0,-0.3);

    \end{tikzpicture}
    \caption{\textbf{Overview of the three-stage post-training pipeline}.}
    \label{fig:post-training-pipeline}
\end{figure}
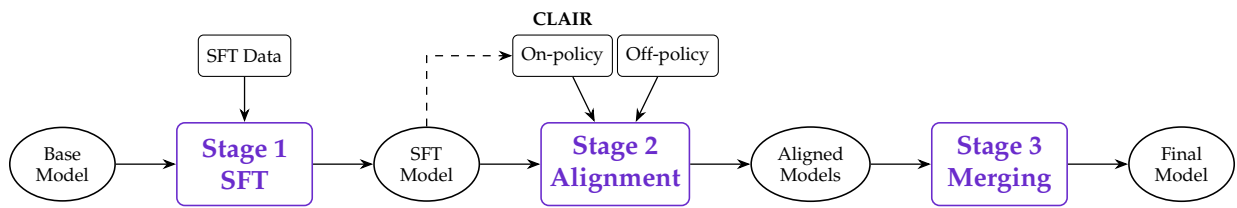

Following the 10-12T token pre-training phase, each LFM2 checkpoint undergoes a three-stage post-training pipeline (Figure~\ref{fig:post-training-pipeline}) with two main objectives. The first goal is teaching the chat template to convert the base model into a conversational assistant that can reliably answer questions, follow instructions, and maintain multi-turn coherence. The second goal is to improve downstream capabilities that are relevant to end users, such as Retrieval-Augmented Generation (RAG) and function calling.

The three-stage pipeline consists of:
\begin{enumerate}
    \item \textbf{Supervised Fine-Tuning (SFT)}: Large general-purpose mixture to provide good capabilities on downstream tasks and teach the chat template.
    \item \textbf{Preference Alignment}: Enabling efficient preference optimization through direct alignment on offline data including on-policy samples generated from the SFT checkpoint.
    \item \textbf{Model Merging}: Systematic evaluation and combination of candidate models to increase robustness and optimize performance.
\end{enumerate}

The following subsections detail data recipes, algorithmic innovations, filtering heuristics, and hyperparameter configurations used in the LFM2 post-training pipeline.

\subsection{Stage 1: Supervised Fine-Tuning}
\label{subsec:sft}

\subsubsection{SFT Data Mixture}

The SFT corpus we use comprises approximately 5.39 million samples for the 350M, 700M, 1.2B, and 2.6B models, and 9.24 million samples for the 8B-A1B model (which includes additional math and code-focused data). The two datasets integrate 67 and 79 carefully curated sources, respectively, spanning open-source datasets, licensed data, and targeted synthetic generations optimized for on-device deployment scenarios, with multilingual samples integrated throughout each category. Table \ref{fig:sft_mixtures} details the composition of the instruction datasets for the LFM2 dense and MoE models. 

\begin{table}[h]
\centering
\begin{subfigure}[t]{0.48\textwidth}
\centering
\vspace{0pt}
\begin{tabular}{lcc}
\toprule
\textbf{Category} & \textbf{Share (\%)} & \textbf{Source} \\
\midrule
General-purpose        & 26.6 & O \\
Instruction following  & 17.1 & P \\
RAG                    & 13.2 & P \\
Real-world chats       & 10.1 & O \\
Tool use               & 10.1 & P \\
Off-policy preferences &  8.2 & P \\
Mathematical reasoning &  7.4 & P,L,O \\
Long context           &  6.7 & P \\
Other                  &  0.6 & O,P \\
                    &      &    \\
\bottomrule
\end{tabular}
\caption{Dense models (350M, 700M, 1.2B, 2.6B)}
\label{subfig:sft_densemixture}
\end{subfigure}
\hfill
\begin{subfigure}[t]{0.48\textwidth}
\centering
\vspace{0pt}
\begin{tabular}{lcc}
\toprule
\textbf{Category} & \textbf{Share (\%)} & \textbf{Source} \\
\midrule
Mathematical reasoning & 26.2 & P,L,O \\
General-purpose        & 16.3 & O \\
Instruction following  & 11.3 & P \\
Code                   & 10.2 & O,P \\
Real-world chats       & 10.0 & O \\
RAG                    &  8.3 & P \\
Tool use               &  7.3 & P \\
Off-policy preferences &  5.3 & P \\
Long context           &  3.2 & P \\
Other                  &  1.9 & O,P \\
\bottomrule
\end{tabular}
\caption{LFM2-8B-A1B}
\label{subfig:sft_moemixture}
\end{subfigure}
\caption{\textbf{SFT data mixture composition for LFM2 models}. Sources: P = proprietary, L = licensed, O = open-source.}
\label{fig:sft_mixtures}
\end{table}

In particular, general-purpose denotes a broad mixture of conversational and task-oriented data (including some code, reasoning, and instruction-like instances) that does not cleanly fall into any single specialized bucket and serves to balance the overall domain coverage. Off-policy preferences correspond to the off-policy component of the preference dataset: we include only the ``chosen'' responses from preference pairs that we ultimately want the model to imitate during SFT, so as to make the supervised data distribution closer to the policy targeted by preference optimization.

\paragraph{Skill-specific mixture design.} We develop the final SFT composition through an iterative process targeting individual capabilities in isolation before combining them into a balanced mixture. This approach allows us to approximate the upper bound performance for each evaluation domain given the training setup. The resulting mixture emphasizes diverse conversational data while maintaining strong performance across technical domains.

\paragraph{Multilingual data.} Rather than treating multilingual capabilities as a separate domain, we systematically integrate multilingual samples across most task categories. The corpus is 80\% English, with the remaining 20\% uniformly distributed across seven languages: Arabic, Chinese, French, German, Japanese, Korean, and Spanish. This approach ensures that the models develop strong multilingual capabilities organically within each skill domain, providing greater diversity and more natural cross-lingual transfer compared to traditional approaches that rely on distinct multilingual datasets. Each category in the mixture contains representative samples in multiple languages, ensuring that specialized capabilities such as mathematical reasoning and tool use are developed multilingually from the ground up.

\paragraph{Data quality pipeline.} We implement a comprehensive multi-stage filtering process to ensure dataset quality and prevent evaluation contamination. As a first-stage quality control step, we perform human evaluation on a subset of each candidate dataset, where the quality of the samples and the diversity of the mixture are assessed to decide which datasets are eligible for downstream use. Only datasets that pass this initial human screen are processed by the subsequent automated stages. 

The filtering process starts with per-dataset quality assessment, where an ensemble of judge LLMs scores the factual accuracy, relevance, and helpfulness of the answers. Quality thresholds vary significantly across sources, with curated proprietary datasets requiring minimal filtering while some open-source datasets underwent more aggressive removal. We then apply validation to eliminate malformed or invalid samples, followed by refusal filtering to remove samples containing refusal patterns or unhelpful responses.

Stylistic filtering uses rule-based methods to eliminate samples with overused phrases, clichés, and undesirable patterns. We perform exact deduplication through direct matching, then apply near-duplicate detection using CMinHash locality-sensitive hashing to catch similar but non-identical content. Semantic deduplication proved most impactful, using sentence embeddings with a high similarity threshold to remove semantically similar content and ensure diverse training signals.

Finally, we perform decontamination through exact n-gram matching against evaluation benchmarks, with additional semantic similarity filtering to catch paraphrased content. The complete pipeline removes a substantial portion of the initial corpus, with semantic deduplication contributing the largest reduction and ensuring non-repetitive, high-quality training data.

\paragraph{Curriculum learning.} We implement a data-driven curriculum learning strategy across the entire dataset to optimize the training progression from simpler to more complex examples. Given the dataset $\mathcal{D} = \{x_i\}_{i=1}^N$ with verifiable rewards and a diverse ensemble of 12 language models $\mathcal{M} = \{m_j\}_{j=1}^{12}$ spanning multiple scales and architectures, we record binary outcomes $r_{ij} \in \{0,1\}$ indicating whether model $m_j$ answered item $x_i$ correctly.

The model ensemble includes: LFM2-350M, LFM2-700M, Hunyuan-500M~\citep{sun2024hunyuanlargeopensourcemoemodel}, LFM2-1.2B, LFM2-2.6B, Phi-4-Mini~\citep{abouelenin2025phi}, Qwen3-4B-Instruct-2507~\citep{yang2025qwen3}, Qwen3-Klear-8B~\citep{Klear-thinking}, ERNIE-4.5-21B-A3B~\citep{ernie2025technicalreport}, Qwen3-30B-A3B-Instruct-2507, Qwen3-30B-A3B-Coder-Instruct, and Qwen3-235B-A22-Instruct-2507. This diverse set spans parameter counts from 350M to 235B and includes both dense and mixture-of-experts architectures.

For each question, we compute the empirical probability of success across the sampled models:
$$
p_i = \frac{1}{J} \sum_{j=1}^J r_{ij}.
$$
High values of $p_i$ correspond to easier questions, while low values identify harder ones. We then train a model to predict and order the probabilities $p_i$ based on prompt features, providing a ranking from easiest to most challenging examples. This allows SFT to proceed in a principled way: starting with items that most models can solve, and gradually introducing those that require deeper reasoning or knowledge.

\subsubsection{Training Protocol}

The training protocol maintains the 32,768 token context length established during mid-training to ensure consistency across the training pipeline. Training is conducted over 3 epochs using a micro-batch size of 1 per GPU with gradient accumulation, resulting in an effective global batch size optimized for each model size.

The learning rate schedule employs cosine decay, starting from a maximum rate of 3×10$^{-5}$ and decaying to a minimum of 1×10$^{-7}$ over the training duration. We incorporate a linear warm-up phase spanning 500 steps, beginning from an initial rate of 1×10$^{-5}$ with a start factor of 1.0. For optimization, we use the AdamW optimizer configured with $\beta_1 = 0.9$, $\beta_2 = 0.95$, and weight decay $\lambda = 0.1$, along with gradient clipping at 1.0 and an epsilon value of 1×10$^{-8}$.

To balance model performance with computational efficiency, we apply 10\% input dropout on embeddings during training only, while avoiding internal dropout layers to preserve inference latency. The training utilizes mixed precision arithmetic with bfloat16 for both activations and gradients, with float32 master weights maintained by the optimizer. The distributed training infrastructure leverages ZeRO-2 sharding with strided activation checkpointing across 8 H100-80GB GPUs per training run to optimize memory usage and training throughput while maintaining fine-grained control over the optimization process.

\subsection{Stage 2: Preference Alignment}
\label{sec:preference_alignment}
We leverage direct alignment methods to achieve a strong trade-off between model capabilities and iteration speed. In particular, we combine a family of length-normalized direct alignment objectives with an offline dataset consisting of both on-policy samples generated from the SFT checkpoint and off-policy reference responses~\citep{rafailov2023direct,meng2024simpo,lambert2024tulu,lisimplemix}.

\subsubsection{Preference Dataset Creation}
\label{sec:preference_data}
The dataset construction focuses on in-distribution samples and combines on-policy generation with skill-specific refinement through larger models. The base dataset consists of open-source as well as proprietary instruction and preference data that constitute approximately 1 million conversations. For each conversation, we sample $N=5$ responses from the selected SFT checkpoint to recover $N+1$ total responses for instruction datasets and $N+2$ total responses for preference datasets. We then score individual responses via an LLM jury, selecting the highest scored sample as the chosen response and the lowest scored sample as the rejected response. Here, we favor on-policy responses over instruction / preference samples in case of a tie to mitigate distribution shift. For instruction following data, we further refine the chosen responses via Contrastive
Learning from AI Revisions (CLAIR)~\citep{d2025anchored}. We apply quantitative score-based filtering to remove low quality preference pairs as well as qualitative filters to mitigate potential undesirable behavior modes. Lastly, we filter out partial samples that exceed the context length. The resulting preference dataset consists of approximately 700,000 conversations.

\subsubsection{Length-Normalized Direct Alignment}
\label{sec:apo_algo}
We implement a family of length-normalized alignment objectives with generalized loss function given by
\begin{align}
\mathcal{L}(\pi_\theta; \pi_{\text{ref}}) &= -\mathbb{E}_{(x,y_w,y_l)\sim\mathcal{D}} \left[ \omega \cdot f \left(\Delta(x,y_w,y_l)-m\right) + \lambda \cdot g\left(\delta(x,y_w,y_l)\right)\right],
\end{align}
evaluated on prompts $x$, chosen responses $y_w$, and rejected responses $y_l$ sampled from dataset $\mathcal{D}$.
Here, we define length-normalized relative $\Delta(x,y_w,y_l)$ as well as absolute $\delta(x,y_w,y_l)$ loss components based on the implicit reward $r_\theta\left(x,y\right)$, such that
\begin{gather}
    \Delta(x,y_w,y_l)=\frac{r_\theta(x,y_w)}{|y_w|} - \frac{r_\theta(x,y_l)}{|y_l|}, \qquad
    \delta(x,y_w,y_l)=\sigma\bigg(\frac{r_\theta(x,y_w)}{|y_w|}\bigg) - \sigma\bigg(\frac{r_\theta(x,y_l)}{|y_l|}\bigg), \\ \nonumber\\ r_\theta\left(x,y\right)=\beta\log\frac{\pi_\theta\left(y|x\right)}{\pi_\text{ref}\left(y|x\right)},
\end{gather}
where $|y|$ denotes token count and $\sigma$ is the sigmoid function. The terms are transformed via functions $f(\cdot)$ and $g(\cdot)$, and composed via weights $\omega$ and $\lambda$. In the above, $\beta$ denotes the standard KL coefficient and $m$ the margin. From the general formulation, we recover length-normalized versions of DPO \{$\omega=1$, $f(x)=\log\sigma(x)$, $m=0$, $\lambda=0$, $g(x)=0$\}~\citep{rafailov2023direct} and APO-zero \{$\omega=0$, $f(x)=0$, $m=0$, $\lambda=1$, $g(x)=x$\}~\citep{d2025anchored} as special cases. We can further construct joint objectives, e.g. \{$\omega=1$, $f(x)=\log\sigma(x)$, $m=0.1$, $\lambda=0.2$, $g(x)=x$\}, to leverage the strengths of both objectives. Notably, the inclusion of the margin $m$ in the relative component aligns with the SimPO objective~\citep{meng2024simpo}, while the absolute component can act as a regularizer to anchor generation probabilities. Reference hyperparameters for the direct alignment runs are provided in Table~\ref{tab:dpo_params}.

\begin{table}[h]
\centering
\begin{tabular}{lc}
\toprule
\textbf{Parameter} & \textbf{Value} \\
\midrule
KL coefficient $\beta$ & $5.0$ \\
Learning rate schedule & Cosine \\
\phantom{Learning rate} max $\eta_{\text{max}}$ & $8\times10^{-7}$ \\
\phantom{Learning rate} min $\eta_{\text{min}}$ & $8\times10^{-8}$ \\
\phantom{Learning rate} warm-up & $0.01$ \\
Global batch size & $2048$ \\
Context length & $1024$ \\
Number of epochs & $2$ \\
\bottomrule
\end{tabular}
\caption{Hyperparameters for direct alignment.}
\label{tab:dpo_params}
\end{table}

\subsection{Stage 3: Model Merging}
\label{subsec:model-merging}

We evaluate LFM2 models on a custom harness with selected public benchmarks, covering knowledge, instruction following, mathematical reasoning, and multilingual domains.

We use multiple parameter-space merging techniques to combine fine-tuned models without requiring additional training. Since merging is computationally inexpensive, we apply different techniques in parallel and evaluate their results to select the best checkpoints. The techniques we test (model soup, task arithmetic, TIES-Merging, DARE, and DELLA) are described in detail in Appendix~\ref{app:model_merging}.

\subsection{Evaluation}
\label{subsec:posttrain_eval}

Small models often fail evaluations because they struggle to output answers in the expected format. Unlike larger models, they have weaker instruction-following and generalization abilities, which creates a gap between their actual answers and what evaluation parsers can extract. To measure their true capabilities more accurately, we build a custom evaluation harness with robust parsing logic. This harness accounts for output quirks specific to models like LFM2, Qwen3, Llama 3.2, Gemma 3, and Granite 4.0, allowing us to extract answers correctly and measure each model's performance fairly. More information about individual benchmarks and their implementations are provided in Appendix~\ref{app:evaluation}.

Tables~\ref{tab:small_models} and~\ref{tab:large_models} present benchmark results for the LFM2 models compared to similarly-sized open-source alternatives. LFM2 models demonstrate competitive performance across multiple evaluation dimensions, particularly excelling in instruction following (IFEval, IFBench, Multi-IF) and mathematical reasoning (GSM8K, GSMPlus, MATH 500, MGSM) tasks relative to model size.

Several notable patterns emerge from the evaluation:

\begin{itemize}
\item \textbf{Knowledge Capabilities:} All LFM2 dense models demonstrate competitive performance in MMLU, MMLU-Pro, and GPQA relative to baselines matched in size. Furthermore, LFM2-8B-A1B achieves a 11.46 point improvement over LFM2-2.6B on MMLU-Pro (37.42\%) while maintaining high inference throughput.

\item \textbf{Instruction Following:} LFM2-1.2B achieves 74.89\% on IFEval, surpassing the 30\% larger Qwen3-1.7B (73.98\%). LFM2-2.6B further improves to 79.56\%, significantly outperforming Llama-3.2-3B (71.43\%) and SmolLM3-3B (72.44\%). Additionally, LFM2-8B-A1B scores 25.85\% on IFBench. These strong results underline the efficacy of refining on-policy instruction following data via CLAIR.

\item \textbf{Mathematical Reasoning:} LFM2-1.2B demonstrates strong multilingual reasoning capabilities, scoring 58.30\% on GSM8K and 55.04\% on MGSM. Despite a modest 7.4\% data allocation for math reasoning, the curriculum-ordered training further scales performance of LFM2-2.6B to 82.41\% on GSM8K and 74.32\% on MGSM. With increased reasoning data allocation, LFM2-8B-A1B yields substantial gains over the 2.6B model, including +10.6 points on MATH 500 and +8 points on MATH Level 5.

\item \textbf{Multilingual Transfer:} Evaluation on MGSM (multilingual GSM8K) and MMMLU (multilingual MMLU) indicates efficient cross-lingual transfer arising from the integrated multilingual training approach. LFM2-1.2B retains 94.4\% of its English GSM8K performance on MGSM (58.30\% vs. 55.04\%) and 84.5\% of its MMLU performance on MMMLU (55.23\% vs. 46.73\%).
\end{itemize}

\clearpage
\begin{table}[tb]
\centering
\small
\setlength{\tabcolsep}{4pt}
\begin{tabular}{lccccccc}
\toprule
\textbf{Benchmark} & \textbf{LFM2-350M} & \textbf{LFM2-700M} & \textbf{LFM2-1.2B} & \textbf{Qwen3-0.6B} & \textbf{Llama-3.2-1B} & \textbf{Gemma-3-1B} & \textbf{Qwen3-1.7B} \\
\midrule
\# Total Params & 0.35B & 0.70B & 1.2B & 0.6B & 1.2B & 1B & 1.7B \\
\# Trained Tokens & 11T & 11T & 11T & 36T & 9T & 2T & 36T \\
\midrule
\multicolumn{8}{c}{\textit{Knowledge}} \\
\midrule
MMLU & 43.43 & 49.90 & 55.23 & 44.93 & 46.60 & 40.08 & 59.11 \\
MMLU-Pro & 15.85 & 18.65 & 19.71 & 27.02 & 19.37 & 13.72 & 43.04 \\
GPQA & 27.46 & 28.48 & 31.47 & 22.14 & 28.84 & 21.07 & 27.72 \\
\midrule
\multicolumn{8}{c}{\textit{Instruction Following}} \\
\midrule
IFEval & 65.12 & 72.23 & 74.89 & 64.24 & 52.39 & 62.90 & 73.98 \\
IFBench & 16.41 & 20.56 & 20.70 & 19.75 & 16.86 & 17.72 & 21.27 \\
Multi-IF & 32.85 & 40.92 & 45.28 & 45.13 & 29.43 & 44.49 & 56.28 \\
\midrule
\multicolumn{8}{c}{\textit{Mathematical Reasoning}} \\
\midrule
GSM8K & 30.10 & 46.40 & 58.30 & 36.47 & 35.71 & 59.59 & 51.40 \\
GSMPlus & 22.16 & 29.99 & 36.09 & 9.10 & 18.57 & 40.16 & 29.62 \\
MATH 500 & 24.20 & 31.80 & 42.40 & 48.60 & 25.00 & 43.40 & 70.00 \\
MATH Lvl 5 & 5.79 & 11.27 & 18.61 & 19.43 & 14.20 & 14.64 & 36.99 \\
\midrule
\multicolumn{8}{c}{\textit{Multilingual}} \\
\midrule
MMMLU & 37.99 & 43.28 & 46.73 & 30.84 & 38.15 & 34.43 & 46.51 \\
MGSM & 29.52 & 45.36 & 55.04 & 41.28 & 29.12 & 43.60 & 66.56 \\
\bottomrule
\end{tabular}
\caption{Performance of tiny language models (350M--2B parameters) across knowledge, instruction following, and mathematical reasoning benchmarks. All results are obtained using our internal evaluation harness, and may differ from other reported scores.}
\label{tab:small_models}
\end{table}

\begin{table}[tb]
\centering
\small
\setlength{\tabcolsep}{4pt}
\begin{tabular}{lccccccc}
\toprule
\textbf{Benchmark} & \textbf{LFM2-8B-A1B} & \textbf{LFM2-2.6B} & \textbf{Llama-3.2-3B} & \textbf{SmolLM3-3B} & \textbf{Gemma-3-4B} & \textbf{Qwen3-4B} & \textbf{Granite-4.0-H} \\
\midrule
\# Total Params & 8.3B & 2.6B & 3.2B & 3.1B & 4B & 4B & 7B \\
\# Active Params & 1.5B & 2.6B & 3.2B & 3.1B & 4B & 4B & 1B \\
\# Trained Tokens & 13T & 11T & 9T & 11T & 4T & 36T & 15T \\
\midrule
\multicolumn{8}{c}{\textit{Knowledge}} \\
\midrule
MMLU & 64.84 & 64.42 & 60.35 & 59.84 & 58.35 & 72.25 & 66.79 \\
MMLU-Pro & 37.42 & 25.96 & 22.25 & 23.90 & 34.76 & 52.31 & 32.03 \\
GPQA & 29.29 & 26.57 & 30.60 & 26.31 & 29.51 & 34.85 & 26.46 \\
\midrule
\multicolumn{8}{c}{\textit{Instruction Following}} \\
\midrule
IFEval & 77.58 & 79.56 & 71.43 & 72.44 & 76.85 & 85.62 & 81.06 \\
IFBench & 25.85 & 22.19 & 20.78 & 17.93 & 23.53 & 30.28 & 18.37 \\
Multi-IF & 58.19 & 60.26 & 50.91 & 58.86 & 66.61 & 75.54 & 52.99 \\
\midrule
\multicolumn{8}{c}{\textit{Mathematical Reasoning}} \\
\midrule
GSM8K & 84.38 & 82.41 & 75.21 & 81.12 & 89.92 & 68.46 & 82.64 \\
GSMPlus & 64.76 & 60.75 & 38.68 & 58.91 & 68.38 & 56.16 & 59.14 \\
MATH 500 & 74.20 & 63.60 & 41.20 & 73.60 & 73.20 & 85.60 & 58.20 \\
MATH Lvl 5 & 62.38 & 54.38 & 24.06 & 51.93 & 52.18 & 73.62 & 36.11 \\
\midrule
\multicolumn{8}{c}{\textit{Multilingual}} \\
\midrule
MMMLU & 55.26 & 55.39 & 47.92 & 50.02 & 50.14 & 60.67 & 56.13 \\
MGSM & 72.40 & 74.32 & 61.68 & 68.72 & 87.28 & 81.76 & 73.68 \\
\bottomrule
\end{tabular}
\caption{Performance of small language models (2B--8B parameters) across knowledge, instruction following, and mathematical reasoning benchmarks. All results are obtained using our internal evaluation harness, and may differ from other reported scores.}
\label{tab:large_models}
\end{table}
\clearpage

\begin{figure}[htb]
  \centering
  \includegraphics[width=1.0\linewidth]{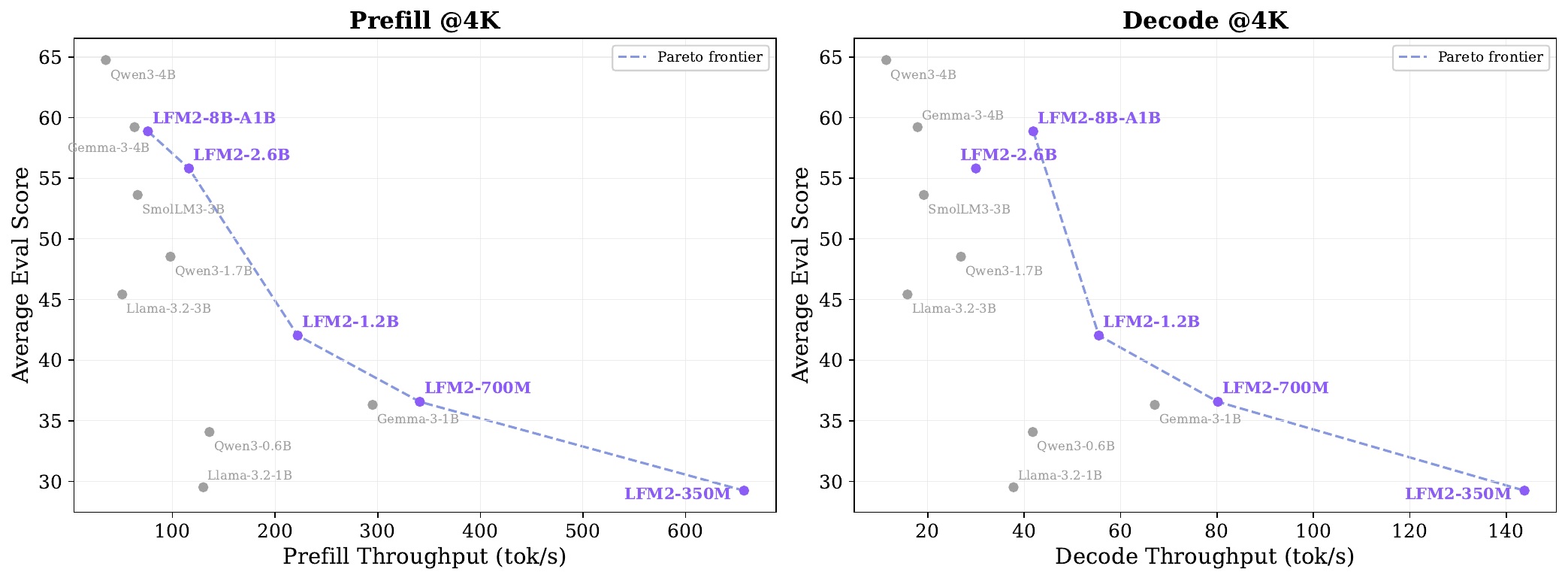}
  \caption{\textbf{Pareto frontier of Average Evaluation Score vs prefill (left) and decode (right) throughput.}
    LFM2 models dominate the Pareto frontier. Each point corresponds to a single model configuration profiled on the Samsung S25 device (Section \ref{subsec:inference}). We chart prefill throughput (tok/s) when processing prompts of length 4k tokens, and decode throughput (tok/s) with a 4k-token prefix. Average Eval Score is computed as the unweighted mean of all evaluation scores reported in Table~\ref{tab:small_models} and Table~\ref{tab:large_models}.
  }
  \label{fig:samsung-s25-pareto}
\end{figure}

The strong performance across diverse benchmarks, paired with superior inference speed, allow LFM2 models to dominate the Pareto frontier of throughput versus average eval score, as shown in Figure~\ref{fig:samsung-s25-pareto}.

\section{Vision-Language LFM2}
\label{sec:vlm}

We extend the LFM2 family to the vision-language setting by augmenting the language backbone with a vision encoder and a lightweight connector, yielding the \textbf{LFM2-VL} models. The overall design and training protocol are similar to other open-source vision-language models (VLMs)~\citep{marafioti2025smolvlm,wang2025internvl3,wang2024qwen2,an2025llava,deitke2024molmo} and, at the same time, preserve the efficiency and deployment characteristics of LFM2.

LFM2-VL is instantiated in three sizes. \textbf{LFM2-VL-450M} combines a Base-86M SigLIP2~\citep{tschannen2025siglip} encoder with an LFM2-350M backbone. \textbf{LFM2-VL-1.6B} and  \textbf{LFM2-VL-3B} combine a So400M SigLIP2 encoder with LFM2-1.2B and LFM2-2.6B backbones, respectively. All variants are designed to support flexible image resolutions, efficient tokenization, and a controllable accuracy vs. latency trade-off for on-device deployment.

In this section, we introduce the LFM2-VL architecture (Section~\ref{sec:vlm_arch}), describe its training recipe (Section~\ref{subsec:vlm_training}), detail the data mixtures used at each training stage (Section~\ref{subsec:vlm-data}), and present evaluation results across a wide range of benchmarks (Section~\ref{subsec:vlm-eval}).

\subsection{Architecture}
\label{sec:vlm_arch}

\begin{figure}
    \centering\includegraphics[width=0.8\linewidth]{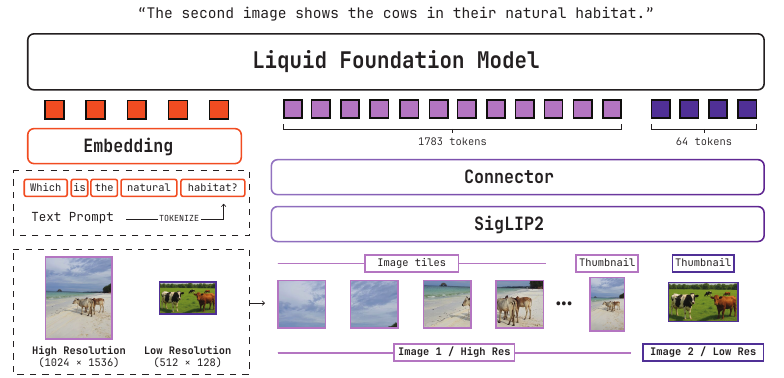}
    \caption{\textbf{LFM2-VL model architecture}. A SigLIP2 image encoder processes images either at native resolution for small inputs or with dynamic tiling for high-resolution inputs. A lightweight connector (PixelUnshuffle + MLP) reduces the number of vision tokens and projects visual embeddings into the LFM2 language token space, enabling unified multimodal processing.}
    \label{fig:vlm_arch}
\end{figure}

All LFM2-VL variants retain the decoder-only LFM2 backbone and attach a SigLIP2 image encoder through a small connector that maps visual features into the language token space as visualized in Figure~\ref{fig:vlm_arch}.

\subsubsection{Vision Encoder and Connector}

For visual inputs, we adopt the NaFlex variant of SigLIP2~\citep{tschannen2025siglip} similar to NaVit~\citep{dehghani2023patch}, which supports variable input resolutions and native aspect ratios. The encoder outputs patch-level embeddings that are projected into language space via a lightweight connector. This connector first applies a PixelUnshuffle~\citep{shi2016real} operation that lowers the number of visual tokens, and then an MLP that maps image embeddings into the LFM2 hidden dimension. The PixelUnshuffle step reduces the number of image tokens by a factor of four, yielding substantial savings in memory and compute with only a slight degradation in downstream multimodal performance in our experiments. For LFM2-VL-1.6B, we use the second-to-last hidden state representation as the image embedding. However, subsequent ablations showed no performance benefit, so for LFM2-VL-450M and LFM2-VL-3B, we use the last hidden state for simplicity.

\subsubsection{Image Preprocessing}

We preprocess images in two regimes depending on their effective resolution: a \emph{single-frame} regime for images up to a resolution of \(512 \cdot  512\) pixels (e.g., \(1024 \times 256\) or \(380 \times 680\)), and a \emph{tiled} regime for larger inputs.

For images in the single-frame regime, we first resize the input so that the SigLIP2 encoder sees a target range of patch counts. The resized height and width are chosen so that both spatial dimensions are divisible by the product of the encoder patch size and the PixelUnshuffle downsampling factor. This ensures alignment with the encoder’s patch grid and removes the need for additional padding prior to PixelUnshuffle. 

For higher-resolution inputs, we switch to dynamic tiling. When the pixel count of an image exceeds a preset threshold, the image is reshaped to the closest supported aspect ratio and partitioned into tiles of size \(512\times 512\) pixels. The number of tiles ranges from 2 to 10, depending on the original resolution. We optionally append a single thumbnail frame, processed through the same single-frame pipeline as above, to provide a coarse global representation of the scene. Tiles are interleaved with special positional tokens, and each multimodal input sequence is wrapped by image start and image end tokens to clearly delineate visual content from text tokens. The optional thumbnail has its own dedicated special token, allowing the model to distinguish global context from local tile information.

Depending on whether the tiling is active and on the size of the image, the number of image tokens ranges from 128 to 256 for untiled images to around 2,800 tokens for heavily tiled inputs. These ranges provide fine-grained control over the image token budget and, consequently, the compute cost during both training and inference.

\subsection{Training Protocol}
\label{subsec:vlm_training}
We train LFM2-VL in three stages, mirroring the language-only pipeline and introducing multimodal alignment at appropriate points. Unless otherwise indicated, the hyperparameters in this section refer to the LFM2-VL-3B configuration; the smaller models follow a similar procedure with appropriately scaled batch sizes and learning rates.

\subsubsection{Stage 1: Connector Pre-training}

The first stage focuses on aligning the visual and textual embedding spaces while keeping the image encoder and language backbone frozen, establishing a mapping from image patch embeddings into the language embedding space before full multimodal training. In this phase, we use a higher learning rate for the connector parameters, a context length of 2,048 tokens, and pad each sample to maximum context length. 

\subsubsection{Stage 2: Multimodal Mid-training}

After language mid-training, we introduce a multimodal mid-training stage that jointly optimizes the language backbone, pre-trained connector, and vision encoder. All components are trainable with a learning rate ratio of 5:5:1 for the text backbone, connector, and encoder, respectively. We found that using a smaller weight decay (1×10$^{-6}$) than in the language training benefits vision training performance. The context length is kept at 32{,}768 tokens, and samples are packed for efficient training. High-resolution tiling is disabled in this phase to maximize sample efficiency, and the learning rate schedule is restarted for approximately 50B additional tokens. In some multimodal mid-training runs, we incorporate additional higher-resolution document OCR data. For these configurations, we set the maximum number of tiles to 4, effectively extending the usable input resolution for more fine-grained OCR while keeping the overall image token budget conservative.

The language mid-training text mixture is reused and augmented with multimodal data (described in Sec \ref{subsec:vlm-data}). We apply a data annealing schedule that starts with 100\% text-only data and, over the first 5\% of steps, linearly decreases its fraction to 30\%, which is then kept fixed for the remainder of training. Ablations showed that the annealing data schedule removes the need for connector pre-training, though we kept the connector pre-training stage as it did not hurt performance. This stage emphasizes robust modality fusion and the introduction of visual knowledge into the model, while maintaining the strong language performance of the mid-trained LFM2 checkpoint.

\subsubsection{Stage 3: Joint Multimodal SFT}

The final stage performs supervised fine-tuning on multimodal instruction-following data, analogous to the language-only SFT pipeline (Section \ref{subsec:sft}). At this point, high-resolution tiling is enabled to expose the model to realistic deployment resolutions. We train several variants on about 50B tokens with diverse text–image ratios, ranging from 5\% to 30\% of total training samples, in order to explore trade-offs between stronger vision and language skills. The context length remains 32{,}768 tokens. This stage aligns the model to downstream use cases such as visual question answering, document understanding, and multi-image reasoning, while simultaneously reinforcing the language-only instruction-following and conversational capabilities.

\subsubsection{Checkpoint Selection and Merging}

For LFM2-VL-3B, we train several candidate runs that vary slightly in training procedure or data composition, for example, through different data sampling ratios, curriculum schedules, OCR-heavy variants, or SFT mixtures with a higher proportion of text-only data. Each candidate is evaluated on an internal multimodal benchmark suite, and the highest-performing checkpoints are combined with a simple linear merge in weight space (as discussed in Section~\ref{subsec:model-merging}). The merged model inherits a balanced combination of text-centric and vision-language capabilities and serves as the final checkpoint.

\subsection{VLM Training Data Mixtures}
\label{subsec:vlm-data}

\paragraph{Connector pre-training mixture.}
The connector pre-training stage uses a captioning corpus designed to provide diverse yet well aligned image-text pairs, with the primary objective of creating an initial alignment signal between the SigLIP2 visual features and the LFM2 language space.

\paragraph{VLM mid-training mixture.}
During multimodal mid-training, we employ a broad mixture of captioning datasets covering diverse scenes and concepts, interleaved image–text documents, OCR-focused data, document visual question answering (VQA), and general VQA datasets. To improve mid-training data quality, we recaption existing open-source image datasets and generate additional VQA samples specifically targeted at visual understanding, including fine-grained reasoning about objects, layouts, and attributes. Some large-scale image datasets used at this stage are filtered for safe-for-work content using automated filters and LLM-based screening; this process removes roughly 0.1\% of samples from the original pools.

We incorporate additional document OCR data with visual augmentations such as blur, noise, and geometric distortions to make the OCR behavior more robust to real-world document scans. We observe that some lower-quality SFT subsets degrade performance when used directly in the SFT stage. To still benefit from their coverage and to familiarize the model with multimodal conversational patterns early in the training stage, we move these datasets to the mid-training stage instead of using them in SFT.

\paragraph{VLM SFT mixture.}
The multimodal SFT mixture spans a wide range of tasks and interaction styles, including both single- and multi-image inputs, single- and multi-turn conversations, and tasks ranging from free-form captioning to multiple-choice questions, short answer tasks, and complex visual reasoning tasks. Around 5\% of the data involves multi-image inputs, and roughly 70\% consists of multi-turn samples, which include both sequences of independent questions and multi-turn dialogues. Most instruction tuning vision datasets are drawn from open-source repositories, with additional synthetic subsets used to strengthen specific downstream behaviors. Notably, we do not employ explicit grounding supervision such as bounding boxes or point coordinates.

\paragraph{VLM multilingual data.}
To strengthen the multilingual capabilities of LFM2-VL-3B, across training stages we integrate vision-language data translated into several target languages, including Arabic, Chinese, French, German, Italian, Japanese, Korean, Portuguese, and Spanish.
The multilingual subset of the captioning and VQA mid-training data enhances the image alignment phase, ensuring that the model sees visually grounded supervision in other languages rather than primarily English. In addition, a portion of the OCR datasets is inherently multilingual, covering a broad range of document types.

Beyond mid-training data, we also translate a variety of English SFT data to target languages, which helps extend multilingual support across the entire task spectrum during the instruction tuning phase. While the translated datasets significantly improve visual capabilities across languages, we acknowledge that further gains will require more culturally contextual and locally sourced data to better capture region-specific visual knowledge.

\subsection{Evaluation}
\label{subsec:vlm-eval}

We evaluate LFM2-VL across a diverse set of multimodal benchmarks, including general visual understanding, visual reasoning, OCR-heavy tasks, math reasoning, and multilingual visual understanding. As shown in Tables~\ref{tab:vl-general} and~\ref{tab:vl-capability}, the LFM2-VL family achieves consistently strong performance relative to similarly sized open-source VLMs. Multilingual scores are based on the average of benchmarks translated by GPT-4.1-mini from English to Arabic, Chinese, French, German, Italian, Japanese, Korean, Portuguese, and Spanish.

\paragraph{Key performance insights.}
\begin{itemize}
    \item \textbf{General Vision Understanding:} Across standard multimodal benchmarks such as MMBench, MMStar, SEEDBench, and RealWorldQA, LFM2-VL models exhibit strong parameter-efficient performance. LFM2-VL-450M outperforms SmolVLM2-500M on MMStar (+3.14) and MMBench (+4.47), while LFM2-VL-1.6B surpasses InternVL3.5-1B on RealWorldQA (+8.63). At the 3B scale, LFM2-VL-3B scores 76.55 on SEEDBench and 71.37 on RealWorldQA, outperforming all baselines, including the larger Qwen2.5-VL-3B model. The models also maintain low object hallucination rate, with LFM2-VL-3B achieving 89.01 on POPE.
\end{itemize}

\begin{table}[tb]
\centering
\small
\begin{tabular}{lcccc}
\toprule
\textbf{Benchmark} & \textbf{LFM2-VL-450M} & \textbf{LFM2-VL-1.6B} & \textbf{SmolVLM2-500M} & \textbf{InternVL3.5-1B} \\
\midrule
\# Total Params  & 0.45B & 1.6B & 0.5B & 1.1B  \\
\midrule
\multicolumn{5}{c}{\textit{General (vision)}} \\
\midrule
MMStar           & 40.87    & 49.87  & 37.73    & 50.27    \\
MMBench (dev)    & 55.76    & 69.16  & 51.29    & 70.79    \\
RealWorldQA      & 52.03    & 65.75  & 51.37    & 57.12    \\
MME              & 1,230 & 1,757 & 1,456 & 1,894 \\
SEEDBench        & 63.62    & 72.00  & 60.34    & 72.67    \\
POPE             & 83.68    & 87.17  & 85.83    & 86.30    \\
\midrule
\multicolumn{5}{c}{\textit{Capability-focused (vision)}} \\
\midrule
MMMU             & 34.44  & 39.67  & 33.89  & 41.89  \\
MathVista        & 45.30  & 51.70  & 34.50  & 53.10  \\
BLINK            & 42.61  & 44.50  & 40.45  & 44.19  \\
InfoVQA (val)    & 44.56  & 58.35  & 24.64  & 60.99  \\
OCRBench         & 657 & 729 & 562 & 790 \\
MM-IFEval        & 33.09  & 46.35  & 11.56  & 36.17  \\
\bottomrule
\end{tabular}
\caption{Performance of VLMs with fewer than 2B parameters on general image understanding and capability-focused benchmarks (multimodal reasoning, mathematical reasoning, multi-image, high resolution input, OCR, and instruction following). All results are obtained using VLMEvalKit~\citep{duan2025vlmevalkitopensourcetoolkitevaluating} and may differ from reported scores of other models.}
\label{tab:vl-general}
\end{table}

\begin{table}[h]
\centering
\small
\begin{tabular}{lccccc}
\toprule
\textbf{Benchmark} & \textbf{LFM2-VL-3B} & \textbf{SmolVLM2-2.2B} & \textbf{InternVL3.5-2B} & \textbf{Qwen3-VL-2B} & \textbf{Qwen2.5-VL-3B} \\
\midrule
\# Total Params & 3.0B & 2.2B & 2.3B & 2.1B & 3.75B \\
\midrule
\multicolumn{6}{c}{\textit{General (vision)}} \\
\midrule
MMStar           & 57.73  & 46.00    & 57.67    & 49.00    & 56.13   \\
MMBench (dev)    & 79.81  & 69.24    & 78.18    & 77.58    & 80.41   \\
RealWorldQA      & 71.37  & 57.50    & 60.78    & 65.75    & 65.23   \\
MME              & 2,051 & 1,793 & 2,129 & 2,045 & 2,163\\
SEEDBench        & 76.55  & 71.30    & 75.41    & 74.08    & 73.88   \\
POPE             & 89.01  & 85.10    & 87.17    & 89.05    & 86.17   \\
\midrule
\multicolumn{6}{c}{\textit{Capability-focused (vision)}} \\
\midrule
MMMU             & 45.33  & 41.60  & 51.78  & 45.78  & 51.67  \\
MathVista        & 62.20  & 51.50  & 61.60  & 63.50  & 62.50  \\
BLINK            & 51.03  & 42.30  & 50.97  & 54.87  & 48.97  \\
InfoVQA (val)    & 67.37  & 37.75  & 69.29  & 71.68  & 76.12  \\
OCRBench         & 822 & 725 & 834 & 864 & 824 \\
MM-IFEval        & 51.83  & 19.42  & 47.31  & 51.35  & 38.62  \\
\midrule
\multicolumn{6}{c}{\textit{Multilingual (vision)}} \\
\midrule
MMBench (dev) & 75.84 & 44.07  & 70.03 & 69.52  & 74.33  \\
MMMB          & 81.52 & 57.82  & 75.86  & 74.11  & 77.60 \\
\midrule
\multicolumn{6}{c}{\textit{Knowledge (language)}} \\
\midrule
MMLU          &  62.70 &  26.89  &  60.89  &  59.87  &  64.92 \\
MMMLU         &  53.37 &  32.16  &  49.32  &  42.20  &  54.23 \\
\bottomrule
\end{tabular}
\caption{Performance of 2--4B VLMs on general image understanding, capability-focused and multilingual vision evaluations, and language benchmarks. All results are obtained using VLMEvalKit~\citep{duan2025vlmevalkitopensourcetoolkitevaluating} or our internal language eval suite, and may differ from reported scores of other models.}
\label{tab:vl-capability}
\end{table}
\clearpage

\begin{itemize}
    \item \textbf{Image Grounded Instruction Following:} The LFM2-VL family performs well in instruction-following tasks. LFM2-VL-450M outperforms SmolVLM-500M by a large margin, demonstrating strong capability even at small scales. LFM2-VL-1.6B achieves 46.35 on MM-IFEval, outperforming InternVL3.5-1B (36.17), while LFM2-VL-3B reaches 51.83, matching or exceeding all open-source models below 4B.
    \item \textbf{Text-Rich and OCR Tasks:} LFM2-VL models demonstrate competitive document understanding and OCR capabilities. The 450M model achieves 657 on OCRBench, substantially higher than SmolVLM2-500M (562), while LFM2-VL-3B scores 822, reaching performance of other similarly sized models.
    \item \textbf{Multilingual Visual Understanding:} LFM2-VL-3B surpasses Qwen2.5-VL-3B, with an average score of 75.84 (+1.51) on translated MMBench and 81.52 (+3.92) on translated MMMB, outperforming all other open-source models under 4B parameters. A detailed breakdown of scores by language are in Table \ref{tab:vl-multilingual} in Appendix \ref{app:vl-multilingual}.
    \item \textbf{Language-Only Tasks:} Despite being optimized for multimodal understanding, LFM2-VL-3B maintains strong language capabilities. The model achieves 62.70 on MMLU and 53.37 on MMMLU, outperforming SmolVLM2-2.2B by large margins (+35.81 on MMLU, +21.21 on MMMLU) and closely matching the performance of larger baselines such as Qwen2.5-VL-3B. These results demonstrate that the training protocol produces strong VLMs without sacrificing language-only performance.
\end{itemize}

These results highlight strong multimodal performance of LFM2-VL models across scales, while also revealing headroom for further gains through reinforcement learning, particularly for more complex multimodal reasoning, such as in the MMMU eval.

\begin{figure}
    \centering\includegraphics[width=1\linewidth]{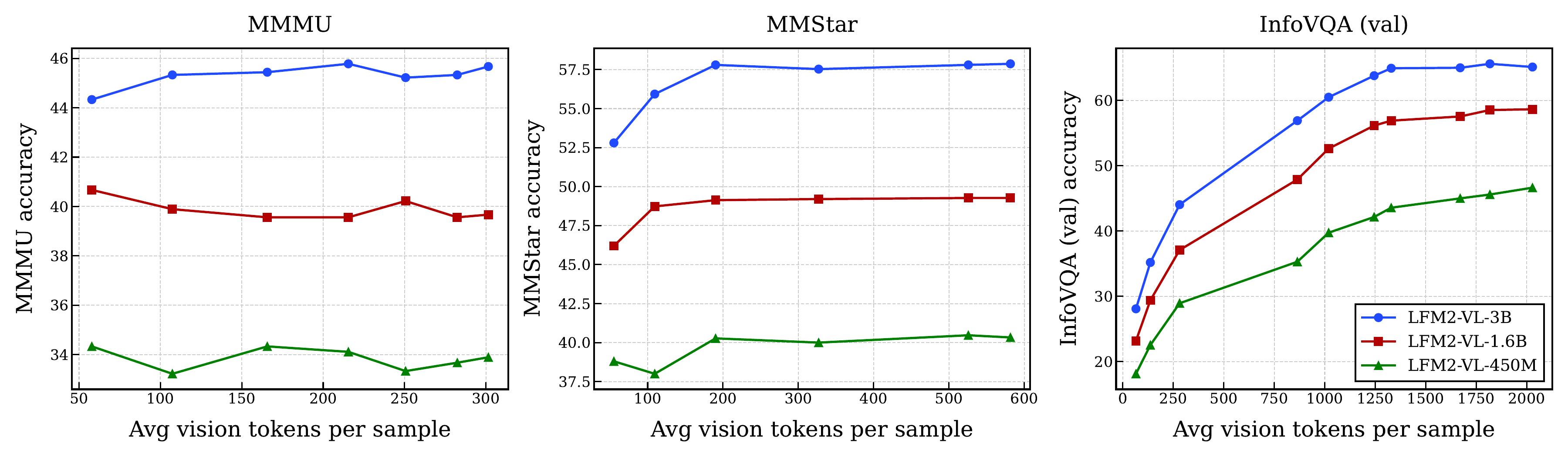}
    \caption{\textbf{Accuracy of LFM2-VL on selected benchmarks under varying vision token budgets.}
We vary the maximum number of vision tokens per image at inference by adjusting the preprocessing pipeline (downsizing single-frame inputs or limiting the number of tiles for high-resolution images), using the token budget as a proxy for on-device latency. Across multimodal reasoning, general vision understanding, and high-resolution perception benchmarks, LFM2-VL retains strong performance under compression, with performance degrading gracefully, especially on high-resolution inputs.}
    \label{fig:vlm-tok-acc}
\end{figure}

\paragraph{Accuracy-vs-latency.} To illustrate the flexibility of LFM2-VL, Figure~\ref{fig:vlm-tok-acc} shows performance as a function of the number of vision tokens at inference time, which serves as a proxy for latency on-device. Reducing the token budget introduces minimal degradation on general vision benchmarks, indicating that the model retains strong vision understanding even under aggressive compression. In contrast, tasks requiring high-resolution perception exhibit more pronounced drops as fewer tokens limit the model’s ability to capture fine-grained detail. These results demonstrate that LFM2-VL offers adaptable performance vs. latency scaling, enabling users to tune computational cost to match visual difficulty and resource constraints.

\section{LFM2-Audio}
\label{sec:audio}
\textbf{LFM2-Audio} extends the LFM2 language model backbone with dedicated components for audio input and output. In contrast to several contemporary audio-language models, LFM2-Audio explicitly separates \emph{continuous} audio input features from \emph{discrete} audio output codes. Audio input is handled via log-mel features and an encoder stack, while audio generation is implemented through a discrete Residual Vector Quantization (RVQ) code path and a lightweight audio detokenizer. This separation preserves a rich representation for speech recognition and understanding while avoiding artifacts introduced by audio tokenization and keeping generation latency low enough for real-time speech-to-speech interaction on edge hardware.

In this section we describe the architecture of \textbf{LFM2-Audio-1.5B} (Section~\ref{sec:audio_arch}), the inference modes it exposes (Section~\ref{sec:audio_inference}), the training procedure and losses (Section~\ref{sec:audio_training}), the data mixtures (Section~\ref{sec:audio_data}), and evaluation results on audio chat and ASR (Automatic Speech Recognition) benchmarks (Section~\ref{sec:audio_eval}).

\subsection{Architecture}
\label{sec:audio_arch}

\begin{figure}
    \centering\includegraphics[width=1\linewidth]{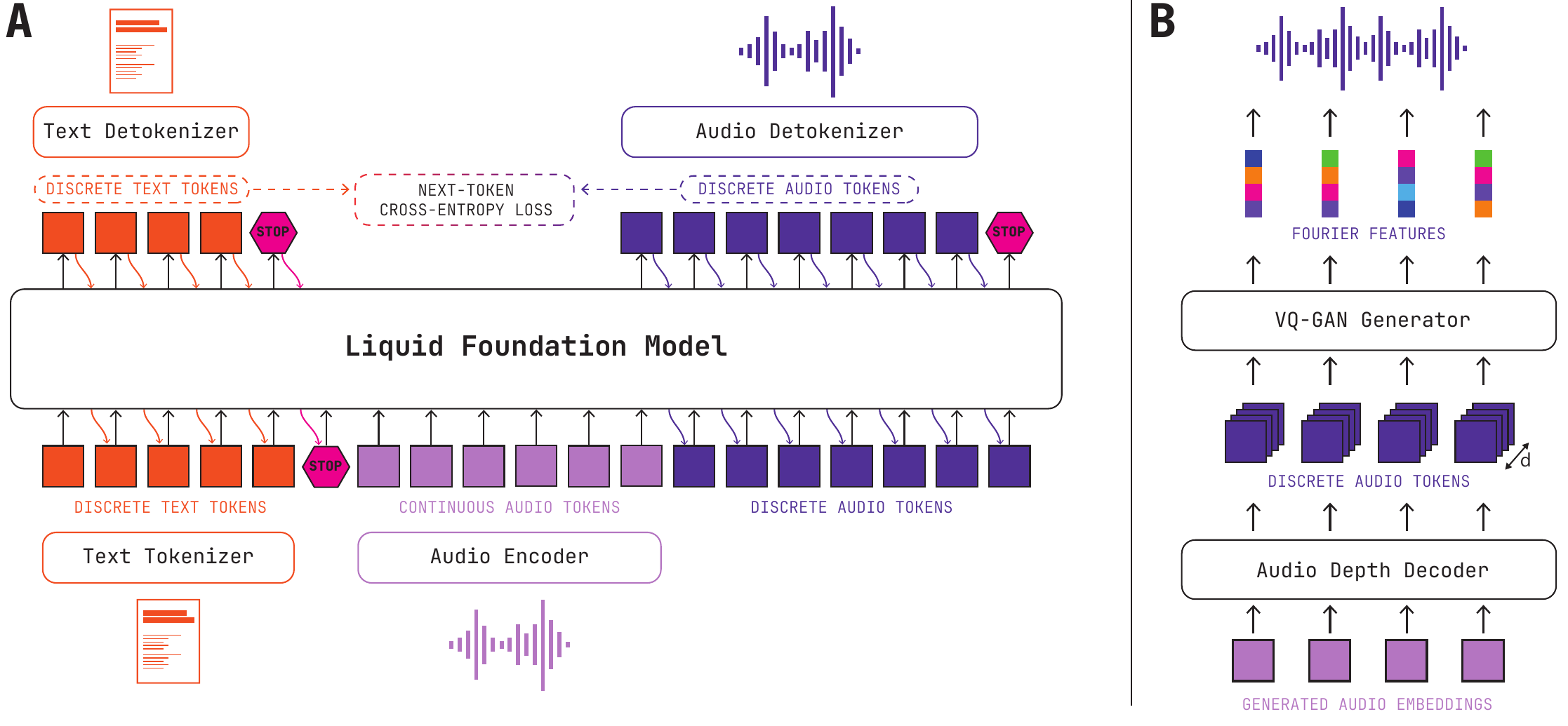}
    \caption{\textbf{LFM2-Audio model architecture}. (A) The LFM2 backbone operates on a single sequence of embeddings that can interleave text and audio tokens, producing next-token predictions.
    (B) The audio generation pipeline. Continuous LFM2 embeddings are turned into discrete code sequences, which are subsequently turned into a waveform by a Fourier based GAN generator.}
    \label{fig:audio-arch}
\end{figure}

LFM2-Audio-1.5B is built around the decoder-only LFM2-1.2B language backbone (Section~\ref{sec:arch}), augmented with an audio encoder on the input side and a discrete audio code path plus detokenizer on the output side.
The main parts of the architecture are sketched in Figure~\ref{fig:audio-arch}.

\subsubsection{Encoder}
The audio encoder produces a sequence of continuous embeddings that can be consumed by the LFM2 backbone in the same way as text token embeddings. Audio inputs are converted to LLM embeddings by an audio encoder followed by an MLP connector. Raw 16~kHz waveforms are first transformed into 128-dimensional log-mel features using a 0.025s window and 0.01s stride. A small stack of strided convolutions performs $8{\times}$ temporal downsampling on the log-mel features, after which 17 FastConformer layers~\citep{rekesh2023fastconformer} process the resulting 512-dimensional continuous features. The MLP connector projects these encoder features to the LFM2-1.2B hidden dimension of 2048. Thus, each resulting embedding corresponds to 0.08~seconds of audio, so the operates at a temporal resolution of 12.5 embeddings per second of input audio.

\subsubsection{Audio Decoder}
For audio output, LFM2-Audio generates \emph{discrete} audio codes rather than continuous waveforms. Specifically, the model predicts 8-codebook Mimi RVQ codes~\citep{defossez2024moshi}, which are later converted to a waveform by a detokenizer. The 8 codebooks consist of one semantic codebook and seven acoustic codebooks, following the Mimi design. Each 8-code frame corresponds to 0.08~seconds of audio, matching the temporal resolution of the encoder.

The combined audio vocabulary contains 2049 tokens per codebook (2048 Mimi tokens plus a special token that marks end-of-audio for that codebook). Once an 8-code frame has been generated, the eight code tokens are embedded and summed to form a single 2048 dimensional embedding, which is fed back into the LFM2 backbone for autoregressive generation. In parallel, the corresponding 8-code frame is streamed to the detokenizer for real-time playback.

To keep inference latency low, we do not ask the LFM2 backbone to predict all eight codes in sequence. Instead, we use a much smaller RQ-Transformer~\citep{lee2022depthformer}, similar to the one used in Moshi~\citep{defossez2024moshi}, as a decoder for code generation. At each time step, the LFM2 backbone produces a single 2048-dimensional output embedding. This embedding is used to condition the RQ-Transformer, which runs for 8 steps to autoregressively generate the 8 codes in the frame (one per codebook). Concretely, when the model is in \emph{audio generation state} (Section~\ref{sec:audio_inference}), each backbone step is followed by 8 RQ-Transformer steps. Because the RQ-Transformer has a much smaller parameter count than the LFM2 backbone (115M vs.~1.2B parameters, respectively), this schedule yields low time-to-first-audio and real-time generation on commodity CPUs compared to other RVQ generation strategies, such as delay patterns \citep{copet2023musicgen}.

\subsubsection{Audio Detokenizer}
Since LFM2-Audio generates Mimi codes, the conversion from codes to a waveform can in principle be done with the Mimi decoder itself or any other compatible model. In practice, we found the Mimi decoder too slow for the target hardware, so we instead train a compact, LFM2-style detokenizer optimized for edge deployment.

The detokenizer closely follows the Vocos architecture~\citep{siuzdak2023vocos}. Its job is to map a sequence of discrete 8-code Mimi frames to a sequence of complex short-time Fourier transform (STFT) coefficients, which are then converted to 24~kHz audio by an inverse STFT. Concretely, the sequence of 8-code frames is first embedded into a sequence of vectors, one embedding per frame. These embeddings are upsampled along the time axis by a factor of 6 using nearest-neighbor upsampling so that the resulting sequence length matches the target number of STFT frames. The upsampled sequence is processed by a small 35M-parameter LFM2 variant with 8 layers (5 gated short convolution blocks and 3 sliding-window attention blocks) and hidden dimension 512. This backbone produces one hidden vector per STFT frame. A final linear projection maps each hidden vector to the complex STFT domain, jointly predicting the log-magnitude and phase for all frequency bins at that frame. Thus, the \emph{network} outputs a sequence of complex STFT coefficients, and the \emph{final} audio waveform is obtained by applying an inverse STFT to these coefficients. This design yields a detokenizer that is fast enough for real-time synthesis on edge hardware while maintaining good perceptual quality.

\subsection{Inference}
\label{sec:audio_inference}
Unlike other speech-to-speech models, LFM2-Audio supports two complementary and distinct inference modes, \emph{interleaved} and \emph{sequential}, that share the same backbone but differ in how they schedule text and audio generation. The choice of mode depends on the downstream task. Interleaved generation is suited to settings that require both non-trivial reasoning and low-latency speech output (e.g., conversational assistants), where the model can emit text and audio in real time. Sequential generation instead produces one modality at a time, with the ability to switch between modalities as needed. This is useful for tasks such as Automatic Speech Recognition (ASR; audio-in, text-out), Text-to-Speech (TTS; text-in, audio-out), or mixed text/audio responses with variable proportions.

We further note that both generation modes are \emph{stateful}, which allows the separation of text and audio vocabularies. The state determines which modality that is being generated at each timestep. In the \emph{text generation state}, the model behaves exactly like the underlying LFM2 base model, using the output embedding to compute next-token probabilities over the text vocabulary. In the \emph{audio generation state}, the model sends the output embedding to the RQ-Transformer and computes next-code probabilities over the audio vocabulary. The different generation modes merely control the state-changing logic for outputs consisting of multiple modalities.

\subsubsection{Interleaved Generation}
In interleaved generation mode, LFM2-Audio generates the assistant response in both text and audio, with a predetermined interleaved pattern. This acts as a kind of chain-of-thought, where the model generates the semantic content in text, leveraging knowledge learned during large-scale text-only pre-training, and subsequently translates the generated text into audio. In our implementation, we generate 6 text tokens followed by 12 audio tokens and repeat until all text tokens are exhausted. When this happens, the model outputs a special \texttt{<|end\_of\_text|>} token, after which it outputs all remaining audio tokens until a special \texttt{<|end\_of\_audio|>} token is generated.

The $6:12$ ratio of text and audio tokens is selected to keep text generation ahead of audio generation and to minimize time-to-first-audio token. Keeping the text-to-audio ratio fixed and separating text and audio vocabularies enables independence from extra control tokens, as is required by other models generating interleaved tokens \citep{li2025baichuan_audio, zeng2024glm4}.

\subsubsection{Sequential Generation}
In sequential generation mode, LFM2-Audio can change its own generation state by generating special control tokens. By default, the model always starts in text generation mode, where only the text vocabulary is used. When a special \texttt{<|start\_of\_audio|>} is encountered, the model switches to the audio generation state and starts using the RQ-Transformer and audio vocabulary to generate outputs. This continues until a special \texttt{<|end\_of\_audio|>} token is generated, after which it switches back to the text state. This design enables generation of responses consisting of both text and audio of arbitrary lengths mixed in arbitrary ways. This mode is suitable for tasks with text-only outputs, such as ASR, since by default the model generates only text. Sequential generation is also used in TTS, by first generating the \texttt{<|start\_of\_audio|>} token, and stopping generation when \texttt{<|end\_of\_audio|>} is generated.

\subsection{Training details}
\label{sec:audio_training}

As a decoder-only model, LFM2-Audio is trained with autoregressive cross-entropy over discrete targets, with separate losses for text and audio tokens.

\paragraph{Token-level loss.}
Let $N$ be the total number of (discrete) tokens in a batch, $V$ the vocabulary size under the current head, and $y_n$ the reference token at position $n$ with corresponding logits $l_{n,i}$. The standard cross-entropy loss is
\begin{align*}
L = -\frac{1}{N} \sum_{n=1}^N \log \frac{\exp(l_{n, y_n})}{\sum_{i=1}^V \exp(l_{n, i})}.
\end{align*}
We compute this loss separately for the text vocabulary, yielding $L_{\mathrm{T}}$, and each of the 8 Mimi codebooks, yielding $L_1,\dots,L_8$.

\paragraph{Codebook weighting.}
We combine the 8 audio codebook losses into a single $L_{\mathrm{A}}$ via a weighted average. Concretely, $L_1$ is assigned weight 100, and other weights receive exponentially decaying weights such that $L_8$ gets a weight 1. This emphasizes semantic and low-index acoustic codes in the loss, which we found empirically to improve perceived audio quality. 

The final loss value is the mean of $L_\mathrm{T}$ and $L_\mathrm{A}$, weighted by the number of tokens of each modality present in a batch. We do not compute losses on continuous audio input embeddings.

\paragraph{Training phases.}
Training proceeds in three phases: alignment, mid-training, and post-training. In the alignment phase, we use ASR data to train the MLP connector, while keeping the LFM2 backbone and encoder frozen. When the loss plateaus, we move on to mid-training. Here, we unfreeze the entire model and train on a large-scale mixture of text and audio datasets. In total, the mid-training phase sees around 100 billion tokens, of which approximately $62\%$ are text tokens, $25\%$ are audio output frames (8-codebook Mimi tokens), and $13\%$ are audio input embeddings (log-mel features).

\paragraph{Detokenizer training.}
The detokenizer is trained separately as a GAN, using both a multi-period discriminator \citep{kong20mpd} and a multi-resolution discriminator \citep{jang21mrd}. The training objective is a combination of log-mel loss (L1), discriminator loss (hinge loss on each frame), and feature matching loss. We train the detokenizer on batches of $16 \times 2.4$ second audio clips for a total of 1,000,000 steps.

\subsection{Audio Training Data Mixtures}
\label{sec:audio_data}

\begin{table}
\centering
\begin{tabular}{lccc}
\toprule
\textbf{Dataset type} & \textbf{\# Text tokens} & \textbf{Input audio (hours)} & \textbf{Output audio (hours)} \\
\midrule
Transcription & 1.5B & 69,163 & -- \\
Text To Speech & 2.6B & -- & 90,826 \\
Language classification & 175M & 4,728 & 851 \\
Audio Chat Instruction Tuning & 4.7B & 72,130 & 234,113 \\
Text Chat Instruction Tuning & 3.7B & -- & -- \\
\midrule
\textbf{Total} & \textbf{$\sim$13B} & \textbf{146,021} & \textbf{325,790} \\
\bottomrule
\end{tabular}
\caption{Dataset composition per epoch across text and audio modalities.}
\label{tab:dataset_composition}
\end{table}

In the mid- and post-training phases, we mix datasets spanning transcription, language classification, text-to-speech, and audio chat instruction tuning across multiple domains. Table~\ref{tab:dataset_composition} presents an overview of the data mixture. The datasets range from established open-source collections such as CommonVoice and LibriSpeech to in-house synthetic datasets generated for specialized audio chat capabilities. For each dataset, we apply augmentation to improve robustness and filtering to ensure quality. Filtering combines rule-based validation (e.g., duration constraints, transcription alignment, and pattern-based filtering) with LLM-judging techniques, where the text or audio quality is evaluated for coherence, correctness, and stylistic naturalness.

The total audio volume used per epoch (approximately 472,000 hours across input and output modalities) is comparable in scale to large-scale speech models such as Whisper, which was trained on around 680,000 hours of audio~\citep{radford2023robust}. In aggregate, the audio data corresponds to roughly 21.5 billion audio embeddings, providing the model with a dense and diverse representation of spoken language and sound across multiple acoustic and conversational domains. This scale ensures that the model is exposed to sufficient variability in speech, style, and content to generalize effectively across downstream audio–language tasks.

A large focus on synthetic conversation data is employed to generate a high volume of multi-turn audio conversations intended to generalize the model for potential downstream use cases such as instruction following, ideation, and general reasoning. These conversations are generated using proprietary data factories that chain together open-source language and audio models to produce natural-sounding exchanges between a user and an audio assistant. Broadly, these pipelines come in two types:
a) pipelines specialized for turning existing text SFT datasets into audio conversations, and
b) pipelines that generate audio conversations following a pre-determined theme using randomly sampled seed specifications.

An important aspect in both pipelines is the introduction of paralinguistic and discourse features characteristic of natural spoken dialogue. This involves the synthetic generation of speech cues and conversational behaviors such as intonation patterns, pauses, filler words (e.g., ``uh," ``you know"), turn-taking signals, and backchannel responses (``mm-hmm," ``right"), making interactions sound human and spontaneous. Additionally, we sample user voices via voice-cloning techniques to teach the model a wide distribution of user voice characteristics (e.g., accent, pitch, timbre, and speaking style). These features are programmatically varied to capture the rhythm, nuance, and acoustic diversity of real conversation, thereby enhancing realism and robustness in downstream audio chat models.

The resulting mixture provides broad coverage of tasks and styles, ensuring the trained model demonstrates robustness across varied speech patterns, dialogue structures, and user voice characteristics.

\subsection{Evaluation}
\label{sec:audio_eval}

We evaluate LFM2-Audio on two sets of tasks: general-purpose speech-to-speech conversational abilities and ASR.

Audio chat capabilities are evaluated using VoiceBench \citep{chen2024voicebench}, which evaluates models on freeform answers (using a GPT4o judge), multiple choice questions, instruction following, and refusals on adversarial prompts. In Table~\ref{tab:audio_VB} we compare LFM2-Audio-1.5B to other similarly sized models.
We note that LFM2-Audio-1.5B remains competitive even against the more than three times larger Qwen2.5-Omni-3B\footnote{Qwen2.5-Omni-3B contains 5B parameters, including a 3B parameter backbone.}. The closest model in total parameter count, Ultravox-v0.5-Llama-3.2-1b, performs similarly but does not support audio generation, a must-have for real-time, low-latency speech-to-speech chatbots.

\begin{table}
\centering
\small
\begin{tabular}{lccccc}
\toprule
\textbf{Benchmark} & \textbf{LFM2-Audio-1.5B} & \textbf{Qwen2.5-Omni-3B} & \textbf{Moshi} & \textbf{Ultravox-v0.5-Llama-3.2-1b} & \textbf{Mini-Omni2} \\
\midrule
\# Total Params & 1.5B & 5B & 7B & 0.7B & 0.6B \\
\midrule
AlpacaEval   & 3.78 & 3.72  & 2.01  & 4.02  & 2.32  \\
CommonEval   & 3.48 & 3.51  & 1.6   & 3.53  & 2.18  \\
WildVoice    & 3.12 & 3.42  & 1.3   & 3.48  & 1.79  \\
SD-QA        & 34.81 & 44.94 & 15.64 & 46.38 & 9.31  \\
MMSU         & 33.99 & 55.29 & 24.04 & 29.99 & 24.27 \\
OBQA         & 45.49 & 76.26 & 25.93 & 33.41 & 26.59 \\
BBH          & 51.2 & 61.3  & 47.4  & 51.4  & 46.4  \\
IFEval       & 30.13 & 32.9  & 10.12 & 43.51 & 11.56 \\
AdvBench     & 98.85 & 88.46 & 44.23 & 97.88 & 57.5  \\
\bottomrule
\end{tabular}
\caption{Performance on VoiceBench. VoiceBench is a collection of 9 speech-in, text-out chatbot tasks. AlpacaEval, CommonEval, and WildVoice are scored on a scale from 0 to 5, whereas the rest are scored on a scale from 0 to 100. Higher scores are better.}
\label{tab:audio_VB}
\end{table}

For ASR, we adopt the framework from the Hugging Face Open ASR Leaderboard \citep{open-asr-leaderboard}, testing ASR performance over 8 different datasets. A comparison between LFM2-Audio and other models can be found in Table~\ref{tab:audio_asr}. Once again, we see that LFM2-Audio-1.5B performs extremely competitively against Qwen2.5-Omni-3B, and approaches the world error rate (WER) numbers of ASR-only models such as Whisper.
Together with the model's strong performance in conversational speech-to-speech interaction, these results make LFM2-Audio well-suited for real-time, low-latency audio-text applications.

\begin{table}
\centering
\small
\begin{tabular}{lccc}
\toprule
\textbf{Benchmark} & \textbf{LFM2-Audio-1.5B} & \textbf{Qwen2.5-Omni-3B} & \textbf{Whisper-large-v3-turbo} \\
\midrule
\# Total Params & 1.5B & 5B & 1.5B \\
\midrule
AMI          & 15.36 & 15.05 & 16.13 \\
Earnings22   & 19.75 & 14.81 & 11.63 \\
Gigaspeech   & 10.63 & 11.76 & 10.14 \\
LS Clean     & 2.03 & 2.14  & 2.10  \\
LS Other     & 4.39 & 4.52  & 4.24  \\
SPGISpeech   & 4.17 & 3.24  & 2.97  \\
Tedlium      & 3.56 & 5.08  & 3.57  \\
Voxpopuli    & 9.93 & 6.59  & 11.87 \\
\bottomrule
\end{tabular}
\caption{Performance on ASR tasks, as measured by Word Error Rate (WER). Lower is better.}
\label{tab:audio_asr}
\end{table}

\section{LFM2-ColBERT}
\label{sec:colbert}

We extend the LFM2 backbone to information retrieval through \textbf{LFM2-ColBERT-350M}, a late interaction retriever optimized for multilingual and cross-lingual semantic search. Late interaction models occupy a unique position in the retrieval landscape: they preserve much of the expressivity of cross-encoders (re-rankers) while retaining the efficiency of bi-encoders, enabling both large-scale retrieval and effective ranking in a single model~\citep{khattab2020colbert}. LFM2-ColBERT-350M demonstrates best-in-class multilingual performance and achieves inference speeds comparable to models with 2.37 times fewer parameters, thanks to the computational efficiency of the LFM2 hybrid architecture.

\subsection{Architecture and Design}

\paragraph{Late interaction paradigm.}
Unlike bi-encoders that compress entire documents and queries into single dense vectors, or cross-encoders that perform expensive token-level interactions, late interaction models compute fine-grained token representations independently for queries and documents, deferring the interaction computation until retrieval time. This design enables pre-computation and efficient indexing of document representations while preserving rich semantic matching capabilities.

\paragraph{Model architecture.}
LFM2-ColBERT-350M builds on the LFM2-350M backbone (16 layers, 1024 hidden dimensions) with an additional 9 task-specific layers, resulting in a 17-layer architecture comprising 10 convolutional blocks, 6 attention blocks, and 1 linear layer. The model processes inputs through the LFM2 transformer to produce contextualized token embeddings, which are then projected to a lower-dimensional space (128 dimensions) optimized for similarity computation:

\begin{equation}
\text{ColBERT}(x) = \mathrm{Linear}(\text{LFM2-Transformer}(x))
\end{equation}

where $x$ represents either a query or document, and the linear layer projects from 1024 to 128 dimensions without bias or additional activation. The total parameter count is 353M, distributed across the hybrid backbone and projection head.

\paragraph{Similarity computation.}
Following the ColBERT framework~\citep{khattab2020colbert}, we compute similarity between a query $q$ and document $d$ using the MaxSim operator:

\begin{equation}
S(q, d) = \sum_{i} \max_{j} \text{sim}(q_i, d_j)
\end{equation}

where $q_i$ and $d_j$ are the $i$-th and $j$-th token embeddings of the query and document respectively, and $\text{sim}(\cdot, \cdot)$ denotes cosine similarity. This operation allows each query token to attend to its most relevant document token, enabling fine-grained semantic matching while maintaining computational efficiency through pre-computed document representations.

\paragraph{Configuration parameters.}
The model supports a maximum context length of 32,768 tokens inherited from the LFM2 backbone. For retrieval tasks, we constrain document inputs to 512 tokens and query inputs to 32 tokens, balancing retrieval quality with computational efficiency. The vocabulary size is 65,536 tokens, using the same byte-level BPE tokenizer as the base LFM2 models.

\subsection{Training Methodology}

We continued to pre-train LFM2-350M to 25T tokens, and LFM2-ColBERT-350M is an extension of this 350M checkpoint. Here we discuss the training method used to convert this LFM2-350M checkpoint to a ColBERT model.

\paragraph{Knowledge distillation framework.}
We train LFM2-ColBERT-350M using knowledge distillation through the PyLate framework~\citep{chaffin2024pylate}, leveraging pre-computed relevance scores from a powerful cross-encoder teacher model. This approach enables the student model to learn fine-grained semantic matching patterns from the teacher's outputs while maintaining the computational efficiency of late interaction retrieval. Knowledge distillation has been shown to produce superior ColBERT models compared to pure contrastive learning approaches~\citep{khattab2020colbert}, as it provides richer training signals through continuous relevance scores rather than binary positive/negative labels.

\paragraph{Training objective.}
We employ the distillation loss implemented in PyLate, which minimizes the mean squared error between student and teacher relevance scores. For a query $q$ with associated documents $\{d_1, \ldots, d_m\}$ and teacher scores $\{s_1^T, \ldots, s_m^T\}$, the loss is:

\begin{equation}
\mathcal{L}_{\text{distill}} = \frac{1}{m} \sum_{i=1}^m \left(s_i^T - S(q, d_i)\right)^2
\end{equation}

where $S(q, d_i)$ is the student's MaxSim score (Equation 2). The PyLate framework applies min-max normalization to teacher scores to ensure numerical stability and consistent gradient magnitudes across different teacher model scales. We train with the normalized scores following best practices from JaColBERTv2.5~\citep{clavie2024jacolbert}.

\paragraph{Optimization details.}
We initialize the transformer backbone from the pre-trained LFM2-350M checkpoint and add 9 randomly initialized layers for the task-specific retrieval head. The model is trained end-to-end using the AdamW optimizer with a learning rate of $3 \times 10^{-5}$, $\beta_1 = 0.9$, $\beta_2 = 0.999$, and weight decay of $0.01$. Training runs for approximately 200,000 steps with a global batch size of 128, distributed across 8 NVIDIA H100 GPUs using BF16 mixed precision. We employ a cosine learning rate schedule with 10\% warmup steps, decaying from the maximum learning rate to $1 \times 10^{-7}$.

During the training process, each instance contains a query and 8 documents with their corresponding teacher scores. Document inputs are truncated to 512 tokens and query inputs to 32 tokens, matching the target deployment constraints.

\subsection{Evaluation}

\paragraph{Benchmarks.}
We evaluate LFM2-ColBERT-350M on NanoBEIR Multilingual~\citep{nanobeir-multilingual}, a condensed version of the BEIR benchmark~\citep{thakur2021beir}, extended to include Japanese and Korean languages. We release this multilingual extension as \texttt{LiquidAI/nanobeir-multilingual-extended} for reproducibility. The benchmark spans 13 diverse retrieval tasks, including fact verification (FEVER, Climate FEVER), question answering (Natural Questions, HotpotQA, FiQA), and entity retrieval (DBPedia, Quora).

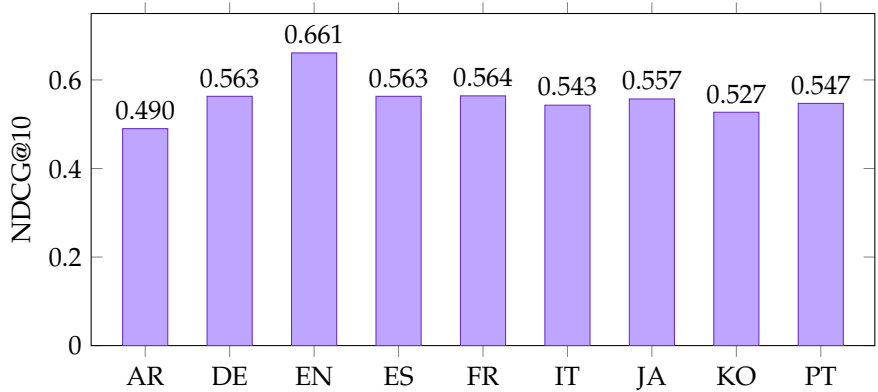
\begin{figure}[h]
\centering
\begin{tikzpicture}
\begin{axis}[
    ybar,
    width=12cm,
    height=6cm,
    bar width=0.6cm,
    ymin=0,
    ymax=0.75,
    ylabel={NDCG@10},
    symbolic x coords={AR, DE, EN, ES, FR, IT, JA, KO, PT},
    xtick=data,
    nodes near coords,
    nodes near coords align={vertical},
    every node near coord/.append style={font=\footnotesize, /pgf/number format/.cd, fixed, fixed zerofill, precision=3},
    enlarge x limits=0.08,
]
\addplot[fill=liquidlavender, draw=liquidpurple] coordinates {
    (AR, 0.490)
    (DE, 0.563)
    (EN, 0.661)
    (ES, 0.563)
    (FR, 0.564)
    (IT, 0.543)
    (JA, 0.557)
    (KO, 0.527)
    (PT, 0.547)
};
\end{axis}
\end{tikzpicture}
\caption{\textbf{Monolingual retrieval performance of LFM2-ColBERT-350M.} Documents and queries are in the same language and measured by NDCG@10 on NanoBEIR across languages.}
\label{fig:colbert_monolingual}
\end{figure}

\paragraph{Monolingual retrieval.}
Figure~\ref{fig:colbert_monolingual} presents NDCG@10 scores across languages when queries and documents are in the same language. LFM2-ColBERT-350M achieves a mean NDCG@10 of 0.661 across all tasks and languages, with particularly strong performance in English (0.661), German (0.563), Spanish (0.563), and French (0.564). The model demonstrates robust multilingual capabilities, maintaining competitive performance even in lower-resource languages like Arabic (0.490), Japanese (0.557), and Korean (0.527).

Compared to GTE-ModernColBERT-v1~\citep{GTE-ModernColBERT}, a similarly-sized baseline with 149M parameters, LFM2-ColBERT-350M shows substantial improvements in multilingual coverage. While the baseline GTE-ModernColBERT-v1 achieves 0.680 NDCG@10 in English, it drops to 0.309 in Arabic, 0.459 in Japanese, and 0.368 in Korean. LFM2-ColBERT-350M maintains more consistent performance across all languages, with the gap between English and lower-resource languages being significantly smaller (0.661 to 0.490 for Arabic, a 26\% reduction vs. 55\% for the baseline).

\paragraph{Cross-lingual retrieval.}
Table~\ref{tab:colbert_crosslingual} presents NDCG@10 scores for all language pairs, where rows represent document languages and columns represent query languages. LFM2-ColBERT-350M is particularly strong in cross-lingual transfer for European languages. For instance, querying in English against French documents achieves 0.551 NDCG@10 (97.7\% of monolingual French performance), and querying in Spanish against Portuguese documents achieves 0.547 (100\% of monolingual Portuguese performance).

\begin{table}[h]
\centering
\footnotesize
\setlength{\tabcolsep}{4pt}
\begin{tabular}{lccccccccc|c}
\toprule
\textbf{Doc / Query} & \textbf{AR} & \textbf{DE} & \textbf{EN} & \textbf{ES} & \textbf{FR} & \textbf{IT} & \textbf{JA} & \textbf{KO} & \textbf{PT} & \textbf{AVG} \\
\midrule
AR & \textbf{0.490} & 0.288 & 0.339 & 0.303 & 0.304 & 0.286 & 0.357 & 0.338 & 0.291 & 33.30 \\
DE & 0.383 & \textbf{0.563} & 0.547 & 0.498 & 0.502 & 0.489 & 0.424 & 0.368 & 0.486 & 47.33 \\
EN & 0.416 & 0.554 & \textbf{0.661} & 0.553 & 0.551 & 0.522 & 0.477 & 0.395 & 0.535 & 51.82 \\
ES & 0.412 & 0.514 & 0.578 & \textbf{0.563} & 0.547 & 0.529 & 0.436 & 0.394 & 0.547 & 50.21 \\
FR & 0.408 & 0.527 & 0.573 & 0.552 & \textbf{0.564} & 0.537 & 0.450 & 0.388 & 0.549 & 50.53 \\
IT & 0.395 & 0.512 & 0.554 & 0.535 & 0.535 & \textbf{0.543} & 0.439 & 0.386 & 0.529 & 49.20 \\
JA & 0.375 & 0.365 & 0.409 & 0.358 & 0.345 & 0.337 & \textbf{0.557} & 0.491 & 0.330 & 39.63 \\
KO & 0.326 & 0.274 & 0.310 & 0.282 & 0.265 & 0.266 & 0.440 & \textbf{0.527} & 0.271 & 32.89 \\
PT & 0.402 & 0.499 & 0.558 & 0.545 & 0.528 & 0.529 & 0.436 & 0.382 & \textbf{0.547} & 49.17 \\
\midrule
\textbf{AVG} & 40.07 & 45.51 & 50.32 & 46.54 & 46.00 & 44.86 & 44.62 & 40.78 & 45.38 & -- \\
\bottomrule
\end{tabular}
\caption{Cross-lingual retrieval performance (NDCG@10) on NanoBEIR for LFM2-ColBERT-350M. Rows represent document language, columns represent query language. Diagonal elements (bold) represent monolingual retrieval. Compare to the results for GTE-ModernColBERT-v1 in Table \ref{tab:gte_colbert_crosslingual}.}
\label{tab:colbert_crosslingual}
\end{table}

\begin{table}[h]
\centering
\footnotesize
\setlength{\tabcolsep}{4pt}
\begin{tabular}{lccccccccc|c}
\toprule
\textbf{Doc / Query} & \textbf{AR} & \textbf{DE} & \textbf{EN} & \textbf{ES} & \textbf{FR} & \textbf{IT} & \textbf{JA} & \textbf{KO} & \textbf{PT} & \textbf{AVG} \\
\midrule
AR & \textbf{0.309} & 0.089 & 0.107 & 0.089 & 0.094 & 0.092 & 0.070 & 0.049 & 0.087 & 10.96 \\
DE & 0.039 & \textbf{0.499} & 0.454 & 0.362 & 0.393 & 0.367 & 0.133 & 0.061 & 0.361 & 29.66 \\
EN & 0.042 & 0.408 & \textbf{0.680} & 0.446 & 0.484 & 0.420 & 0.167 & 0.073 & 0.438 & 35.09 \\
ES & 0.044 & 0.360 & 0.485 & \textbf{0.525} & 0.465 & 0.437 & 0.149 & 0.061 & 0.487 & 33.48 \\
FR & 0.044 & 0.381 & 0.505 & 0.455 & \textbf{0.546} & 0.428 & 0.136 & 0.057 & 0.467 & 33.54 \\
IT & 0.043 & 0.369 & 0.449 & 0.446 & 0.451 & \textbf{0.516} & 0.143 & 0.054 & 0.448 & 32.43 \\
JA & 0.031 & 0.169 & 0.250 & 0.172 & 0.177 & 0.169 & \textbf{0.459} & 0.059 & 0.165 & 18.34 \\
KO & 0.030 & 0.134 & 0.169 & 0.127 & 0.133 & 0.125 & 0.090 & \textbf{0.368} & 0.124 & 14.44 \\
PT & 0.043 & 0.368 & 0.479 & 0.492 & 0.467 & 0.448 & 0.138 & 0.062 & \textbf{0.530} & 33.63 \\
\midrule
\textbf{AVG} & 6.94 & 30.86 & 39.76 & 34.60 & 35.67 & 33.36 & 16.50 & 9.38 & 34.52 & -- \\
\bottomrule
\end{tabular}
\caption{Cross-lingual retrieval performance (NDCG@10) on NanoBEIR for GTE-ModernColBERT-v1. Rows represent document language, columns represent query language. Diagonal elements (bold) represent monolingual retrieval. Compare to the results for LFM2-ColBERT-350M in Table \ref{tab:colbert_crosslingual}.}
\label{tab:gte_colbert_crosslingual}
\end{table}

Table~\ref{tab:gte_colbert_crosslingual} shows that GTE-ModernColBERT-v1 suffers from more degradation in cross-lingual scenarios. Querying in German against English documents yields only 0.408 NDCG@10 (60\% of monolingual English performance), and most cross-lingual combinations involving Arabic, Japanese, or Korean fall below 0.200.

\paragraph{NanoBEIR task breakdown.}
Table~\ref{tab:colbert_nanobeir} presents detailed results on individual NanoBEIR tasks, showing consistent performance across diverse retrieval scenarios. The model excels in fact verification tasks (FEVER: 0.949 NDCG@10, HotpotQA: 0.895 NDCG@10) and maintains strong performance in question answering (Natural Questions: 0.746 NDCG@10) and domain-specific retrieval (SciFact: 0.804 NDCG@10, FiQA: 0.591 NDCG@10).

\begin{table}[h]
\centering
\small
\begin{tabular}{lcccccc}
\toprule
\textbf{Dataset} & \textbf{NDCG@10} & \textbf{MRR@10} & \textbf{MAP@100} & \textbf{Acc@1} & \textbf{Acc@10} & \textbf{Recall@10} \\
\midrule
NanoFEVER & 0.949 & 0.967 & 0.940 & 0.96 & 0.98 & 0.960 \\
NanoHotpotQA & 0.895 & 0.954 & 0.845 & 0.92 & 1.00 & 0.940 \\
NanoQuoraRetrieval & 0.882 & 0.863 & 0.843 & 0.80 & 1.00 & 0.979 \\
NanoSciFact & 0.804 & 0.771 & 0.771 & 0.70 & 0.92 & 0.910 \\
NanoNQ & 0.746 & 0.732 & 0.708 & 0.66 & 0.88 & 0.850 \\
NanoDBPedia & 0.714 & 0.898 & 0.575 & 0.86 & 0.98 & 0.418 \\
NanoMSMARCO & 0.686 & 0.644 & 0.656 & 0.58 & 0.82 & 0.820 \\
NanoTouche2020 & 0.630 & 0.889 & 0.462 & 0.80 & 1.00 & 0.347 \\
NanoFiQA2018 & 0.591 & 0.663 & 0.533 & 0.56 & 0.82 & 0.636 \\
NanoSCIDOCS & 0.384 & 0.613 & 0.297 & 0.50 & 0.86 & 0.381 \\
NanoArguAna & 0.551 & 0.449 & 0.451 & 0.28 & 0.88 & 0.880 \\
NanoClimateFEVER & 0.387 & 0.506 & 0.313 & 0.40 & 0.80 & 0.459 \\
NanoNFCorpus & 0.377 & 0.566 & 0.184 & 0.50 & 0.70 & 0.152 \\
\midrule
\textbf{Mean} & 0.661 & 0.732 & 0.583 & 0.655 & 0.895 & 0.672 \\
\bottomrule
\end{tabular}
\caption{Detailed results on NanoBEIR tasks for LFM2-ColBERT-350M. All metrics are reported on English queries and documents.}
\label{tab:colbert_nanobeir}
\end{table}

\paragraph{Multilingual capability inheritance.}
Despite training exclusively on English data, the resulting model demonstrates strong multilingual retrieval capabilities (Figure~\ref{fig:colbert_monolingual}) and robust cross-lingual transfer (Table~\ref{tab:colbert_crosslingual}). During pre-training, the LFM2-350M checkpoint we use here was exposed to text in several languages, including English, Arabic, Chinese, French, German, Japanese, Korean, Spanish, across 25 trillion tokens. Afterwards, during the knowledge distillation phase, the model was only exposed to English tokens. This transfer mechanism enables zero-shot multilingual retrieval without requiring parallel training data in target languages.

The effectiveness of this approach depends on the similarity between the pre-training multilingual data distribution and the target retrieval tasks. The results show stronger cross-lingual transfer for European languages (0.50+ NDCG@10 cross-lingually) compared to more distant language pairs like Arabic-Japanese (0.375 NDCG@10), likely reflecting both linguistic distance and the relative volume of each language in the pre-training corpus. Future work could investigate whether targeted multilingual fine-tuning on translated English data or parallel retrieval corpora further improves cross-lingual performance, particularly for low-resource language pairs.

\subsection{Inference Performance}

\paragraph{Experimental setup.}
rWe benchmark encoding throughput on NVIDIA H100 GPUs across batch sizes from 1 to 64 for query encoding and 1 to 32 for document encoding. All measurements use identical quantization (BF16), tokenizer settings, and hardware configurations. Query inputs are constrained to 32 tokens following realistic patterns from MS MARCO~\citep{msmarco} and Natural Questions~\citep{kwiatkowski-etal-2019-natural} datasets. Document inputs use 512 tokens sampled from diverse domains to capture realistic length distributions. We report throughput in sequences per second, averaged over 1000 warmup iterations followed by 10,000 timed iterations.

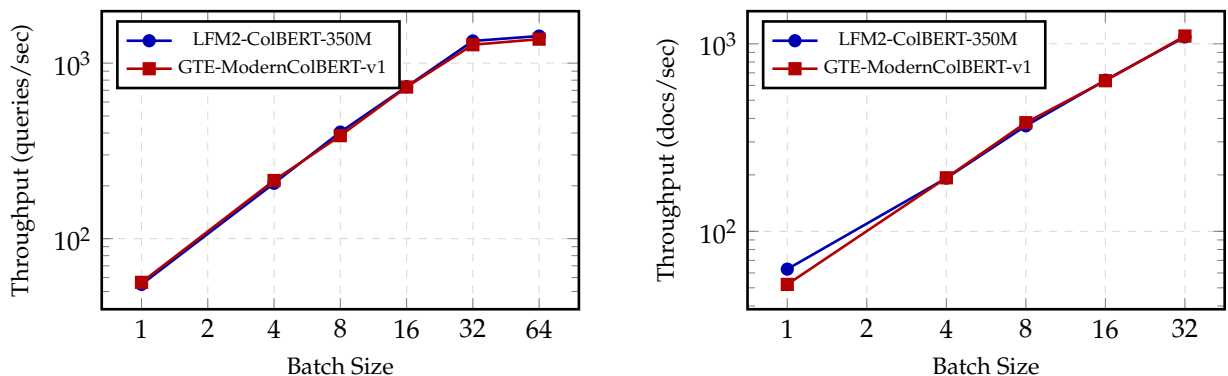
\begin{figure}[htbp]
    \centering
    \begin{subfigure}[b]{0.48\textwidth}
        \centering
        \begin{tikzpicture}
            \begin{axis}[
                width=\textwidth,
                height=0.7\textwidth,
                xlabel={Batch Size},
                ylabel={Throughput (queries/sec)},
                label style={font=\small},
                tick label style={font=\footnotesize},
                legend style={font=\scriptsize},
                grid=major,
                grid style={dashed, gray!30},
                legend pos=north west,
                mark size=2pt,
                line width=1pt,
                xmode=log,
                ymode=log,
                log basis x=2,
                xtick={1,2,4,8,16,32,64},
                xticklabels={1,2,4,8,16,32,64},
                ymajorgrids=true,
                xmajorgrids=true,
            ]
            
            \addplot[
                color=blue!70!black,
                mark=*,
            ] coordinates {
                (1, 54.86)
                (4, 206.67)
                (8, 404.67)
                (16, 734.24)
                (32, 1337.51)
                (64, 1427.14)
            };
            \addlegendentry{LFM2-ColBERT-350M}
            
            \addplot[
                color=red!70!black,
                mark=square*,
            ] coordinates {
                (1, 56.36)
                (4, 214.46)
                (8, 386.07)
                (16, 730.98)
                (32, 1272.87)
                (64, 1369.55)
            };
            \addlegendentry{GTE-ModernColBERT-v1}
            
            \end{axis}
        \end{tikzpicture}
        \caption{Query encoding throughput (queries/sec). Query length fixed at 32 tokens.}
        \label{fig:colbert_query_throughput}
    \end{subfigure}
    \hfill
    \begin{subfigure}[b]{0.48\textwidth}
        \centering
        \begin{tikzpicture}
            \begin{axis}[
                width=\textwidth,
                height=0.7\textwidth,
                xlabel={Batch Size},
                ylabel={Throughput (docs/sec)},
                label style={font=\small},
                tick label style={font=\footnotesize},
                legend style={font=\scriptsize},
                grid=major,
                grid style={dashed, gray!30},
                legend pos=north west,
                mark size=2pt,
                line width=1pt,
                xmode=log,
                ymode=log,
                log basis x=2,
                xtick={1,2,4,8,16,32},
                xticklabels={1,2,4,8,16,32},
                ymajorgrids=true,
                xmajorgrids=true,
            ]
            
            \addplot[
                color=blue!70!black,
                mark=*,
            ] coordinates {
                (1, 62.88)
                (4, 192.31)
                (8, 365.37)
                (16, 640.05)
                (32, 1088.16)
            };
            \addlegendentry{LFM2-ColBERT-350M}
            
            \addplot[
                color=red!70!black,
                mark=square*,
            ] coordinates {
                (1, 52.13)
                (4, 193.00)
                (8, 379.79)
                (16, 635.57)
                (32, 1100.45)
            };
            \addlegendentry{GTE-ModernColBERT-v1}
            
            \end{axis}
        \end{tikzpicture}
        \caption{Document encoding throughput (docs/sec). Document length fixed at 512 tokens.}
        \label{fig:colbert_doc_throughput}
    \end{subfigure}
    \caption{\textbf{Encoding throughput on NVIDIA H100 GPU across batch sizes}. LFM2-ColBERT-350M (353M parameters) matches the throughput of GTE-ModernColBERT-v1 (149M parameters).}
    \label{fig:colbert_throughput}
\end{figure}

\paragraph{Query encoding results.}
Figure~\ref{fig:colbert_query_throughput} presents query encoding throughput across batch sizes. At batch size 1, both models achieve approximately 60 queries/second. As batch size increases, LFM2-ColBERT-350M demonstrates efficient scaling, reaching 1,350 queries/second at batch size 32 and 1,420 queries/second at batch size 64. GTE-ModernColBERT-v1 achieves 1,300 and 1,370 queries/second at the same batch sizes, respectively. LFM2-ColBERT-350M thus maintains a 3.8\% throughput advantage at batch size 32 and a 3.6\% advantage at batch size 64, despite being 2.37 times larger. The throughput curves show saturation around batch size 32, indicating optimal GPU utilization where memory bandwidth becomes the limiting factor rather than compute capacity.

\paragraph{Document encoding results.}
Figure~\ref{fig:colbert_doc_throughput} presents document encoding throughput. The performance curves are nearly indistinguishable, with both models achieving approximately 60 documents/second at batch size 1 and approximately 1,100 documents/second at batch size 32.

\section{Related Work}
\label{sec:related_work}
Here, we further situate LFM2 relative to three threads: (i) modern open foundation model releases, (ii) architectural choices for efficiency, (iii) training methods for small models.

\subsection{Modern Open Weight Foundation Model Families}
Modern autoregressive language modeling generally follows the GPT-3 paradigm of large decoder-only Transformers trained on diverse web text \citep{brown2020language}, with capability gains predicted by scaling laws over data, compute, and parameters \citep{kaplan2020scaling,hoffmann2022training}. LFM2 adapts this lineage to an \emph{edge-first} objective.

Recent general-purpose families~\citep{liu2024deepseek, yang2025qwen3,grattafiori2024llama, team2025gemma} optimize primarily for broad capability and efficient datacenter serving (high throughput, large batches, accelerator-rich environments), though sometimes also include smaller model checkpoints. Other notable model families that have been useful for open science include the OLMo series of models~\citep{groeneveld2024olmo, olmo20242,muennighoff2024olmoe}. By contrast, LFM2 is edge-first and addresses an efficiency and quality gap for on-device models by co-designing architecture and training procedures specifically for on-device constraints while preserving the generalist capabilities that make foundation models valuable.

\subsection{Alternative Architectures for Sequence Modeling}
\label{subsec:rel_work_alt_seq}
\paragraph{Liquid Time-Constant Networks}
Our work draws inspiration from Liquid Time-Constant (LTC) networks which introduce continuous-time neural dynamics with input-conditioned time constants for sequence modeling~\citep{hasani2021liquid, lechner2020neural, hasani2022closed}. LTC networks demonstrate that neural ODEs with adaptive time constants can achieve superior performance on time-series tasks while maintaining interpretability and computational efficiency \citep{chahine2023robust,hasani2020natural,vorbach2021causal,lechner2020gershgorin, lechner2019designing}. The liquid formulation's ability to dynamically adjust its effective temporal receptive field based on input characteristics inspired LFM2's adaptive processing of local and global context.

\paragraph{State space models, linear attention and convolutions}
Another line of work that influences the LFM2 design is the development of efficient softmax attention alternatives that take the form of linear state space models (SSMs), linear attention, or convolutions.  The S4~\citep{gu2022efficiently} model introduced a linear state space parameterization enabling efficient long-range sequence modeling by using a long convolution formulation for parallel processing and a fixed-state size recurrent formulation for autoregressive decoding. 
Further algorithmic improvements to this class of models were achieved by directly leveraging an alternative frequency-domain representation, the z-domain rational transfer function, which reduces the entire linear SSM parallel inference algorithm to a small number of FFT operations over the input and parameters \citep{parnichkun2024state}.
However, a limitation of these approaches was that it required a linear time-invariant formulation, preventing the ability to use input-dependent dynamics.  Liquid-S4~\citep{hasani2023liquid} showed how a particular input-dependent state transition could be incorporated into S4 while maintaining the convolutional computation path, leading to improved performance. Meanwhile, S5~\citep{smith2023simplified, smith2023convolutional} unlocked the ability to use time-varying SSMs by showing that fully recurrent SSMs could be computed using efficient parallel scans and diagonalized dynamics, obviating the need for the convolution formulation and LTI restriction. This SSM line of work led to the S6 formulation and Mamba block~\citep{gu2024mamba}, which introduced a hardware-aware parallel scan implementation that enabled the ability to use input-dependent SSMs with extremely large states. 

In parallel, linear attention variants~\citep{katharopoulos2020transformers, choromanski2021rethinking} reparameterize or approximate the softmax activation and often take advantage of the associativity of matrix multiplications to admit more efficient parallel processing and fixed-state size recurrences.  More recent versions improve on the gating and parameterization of the update rules to increase expressivity~\citep{yang2024gated, dao2024transformers, yang2024parallelizing, yang2025gated}.

Attempts to perform language modeling with efficient SSMs or linear attention led to more complex computational blocks to improve expressivity. The Gated State Space (GSS)~\citep{mehta2023long} model introduced the idea of adding additional multiplicative gating around an SSM. The H3 model~\citep{fu2023hungry} augmented this design by adding an additional short-range shift SSM (equivalent to a short-range convolution) along with a long-range SSM and extra gating in-between. Hyena~\citep{poli2023hyena} replaced the long-range SSM with an implicitly parametrized long-range convolution~\citep{sitzmann2020implicitneuralrepresentationsperiodic, romero2022ckconv} and replaced the shift SSM with a short convolution. While the long-range convolution prevents efficient autoregressive decoding~\citep{massaroli2023laughing}, the idea of augmenting SSM or linear attention blocks with a short convolution has remained in various other designs~\citep{gu2024mamba, dao2024transformers, yang2024parallelizing}.

\paragraph{Hybrid architectures}
While some studies have shown the ability of sub-quadratic architectures (SSMs, linear attention, long convolutions)  to match softmax attention in perplexity and some public benchmarks, a growing body of work indicates that \emph{pure} sub-quadratic architectures degrade on long-range in-context abilities~\citep{park2024can}, such as long-range retrieval~\citep{wen2025rnns, blouir2024birdie}, multi-query associative recall~\citep{arora2024zoology, arora2024simple}, and copying~\citep{jelassi2024repeat}. In addition, they exhibit structurally bounded memory capacity and often much lower effective state size on memory-intensive tasks compared to softmax attention~\citep{parnichkun2025quantifying}. These types of core skills are critical for LLMs that need to be able to generate coherent text, follow directions, aggregate information, and respond
accurately to multiple queries.

A typical approach to address these weaknesses has been to formulate hybrid models that interleave sub-quadratic layers with global attention layers. While previous works such as Longformer~\citep{beltagy2020longformer} Big Bird~\citep{zaheer2020big}, and GPT-3~\citep{brown2020language} explored hybrid architectures of various softmax attention variants (e.g., sliding window attention, sparse attention), to our knowledge, the GSS work was the first to propose interleaving softmax attention with sub-quadratic architectures~\citep{mehta2023long} to overcome performance deficits of pure sub-quadratic models. Other works followed suit~\citep{fu2023hungry, park2024can} to improve observed skill degradations. Hybrid strategies with varying amounts of attention have become standard for released alternative architectures~\citep{lieber2024jamba, waleffe2024empirical, chen2025minimax, team2025kimi}.

LFM2 follows this general principle of hybridization, but adopts a minimal hybrid tuned for devices with the hardware-in-the-loop search: gated short convolutions handle most local mixing, and only a minority of GQA blocks provide global context processing. We find that for on-device settings under identical efficiency budgets, further augmenting these stacks with SSMs, linear attention, or additional convolution operations did not improve quality (\ref{subsec:arch_opt}). We note that there are connections between these findings and the recently proposed Canon layers~\citep{allen-zhu2025physics}.

\subsection{Efficient Training for Small Models}
\label{subsec:rel_work_distill}
The knowledge distillation approach introduced in Section \ref{subsec:KD} extends classical distillation~\citep{hinton2015distilling}  through a decoupled Top-K objective that addresses support mismatch issues when distilling from larger models. This approach is closest to the \emph{ghost token} device discussed in concurrent Sparse Logit Sampling (SLS) work~\citep{anshumann2025sparse}, where the teacher’s tail probability is aggregated into one token and KL is computed on the augmented support. By the KL chain rule, at $\tau=1$ this is exactly a binary membership KL plus a conditional Top-K KL as in our approach. Our contribution makes this decomposition explicit for Top-K storage and introduces temperature \emph{only} inside the conditional Top-K term, which avoids the support mismatch that arises when tempering is applied over the augmented support. We note that our objective is deterministic and complementary to the sampling-based path in SLS. More broadly, this decomposition echoes the spirit of Decoupled Knowledge Distillation (DKD)~\citep{zhao2022decoupled}, but unlike DKD we target the Top-K setting where tail logits are unavailable, and storage is the primary constraint.

\section{Conclusion}
\label{sec:conclusion}
We presented LFM2, the second generation of Liquid Foundation Models, designed for on-device deployment under tight latency and memory constraints. The model family includes dense variants from 350M to 2.6B parameters and an 8.3B MoE model with 1.5B active parameters, all supporting 32K context length. We extend this backbone to multimodal and retrieval domains with LFM2-VL for vision–language tasks, LFM2-Audio for speech, and LFM2-ColBERT-350M for late-interaction retrieval.

\subsection{Key Contributions}
LFM2 combines architectural design, training, and deployment choices around an edge-first objective.

\begin{itemize}
  \item \textbf{Minimal hybrid backbone.} A hardware-in-the-loop search over architectures and layouts yields a compact hybrid that uses gated short convolutions for most layers and a small number of GQA blocks for global context. Under identical quantization and runtimes, this design delivers up to 2$\times$ faster prefill and decode on CPUs compared to similarly sized attention-heavy baselines, while matching or improving benchmark accuracy. 
  
  \item \textbf{Efficient pre-training} A tempered, decoupled Top-K knowledge distillation objective reduces storage and bandwidth when distilling from a larger teacher while avoiding support mismatch and stabilizing losses.

  \item \textbf{Post-training for edge workflows} A three-stage post-training recipe including supervised fine-tuning, length-normalized direct preference alignment, and parameter-space model merging that improves instruction following, RAG, function calling, and multilingual robustness for small models.

  \item \textbf{Multimodal and retrieval extensions} LFM2-VL adds a SigLIP2-based vision encoder with dynamic tiling and token reduction, enabling a flexible resolution–latency trade-off. LFM2-Audio separates continuous audio input from discrete audio output, enabling real-time speech-to-speech capabilities with an efficient model. LFM2-ColBERT-350M adapts the hybrid backbone to late-interaction retrieval, achieving strong multilingual and cross-lingual performance with throughput comparable to much smaller attention-only baselines.

\end{itemize}

Together, these components show that an architecture and training procedure co-designed with device constraints can deliver broadly capable models that remain practical to deploy on commodity edge hardware.

\subsection{Limitations and Future Work}
While LFM2 achieves strong results for its size class, several limitations and opportunities for future work remain.

\paragraph{Hardware \& deployment coverage.}

The deployment recipes and architecture search considered here are tuned for batch size of 1, low-latency inference on a small set of CPU and mobile SoC configurations (Snapdragon-class phones and Ryzen-class laptops) using specific quantization schemes (8da4w in ExecuTorch and Q4\_0 in llama.cpp). LFM2 also runs competitively on modern NPUs and GPUs, but these accelerators were not central to the hardware-in-the-loop search, and we do not claim the resulting architectures or quantization schemes are optimal for large-batch server settings or any particular accelerator family. The trade-off between specializing model variants for specific hardware targets versus maintaining a single cross-platform backbone remains open. As edge runtimes, compiler stacks, and NPU/GPU kernel libraries continue to mature, a broader region of the model and quantization design space may become attractive. Revisiting LFM2’s architecture and deployment recipes under improved edge infrastructure is therefore an important direction for future work.

\paragraph{Capacity and task coverage.} 

As small-scale models, LFM2 variants are inherently capacity-limited compared to frontier-scale systems: they cannot match the breadth and depth of capabilities across all open-ended reasoning, knowledge-intensive, and highly compositional tasks. In practice, we observe that LFM2 performs best on workloads that align with its design targets: short to medium context interactions, task-oriented applications, and edge deployments with tight latency and memory budgets. We view LFM2 as a step in an ongoing line of work to expand the capabilities of small models through improved data, objectives, and architectures. We are continuing to explore training and deployment techniques that narrow the gap with larger models for targeted use cases.

\paragraph{Multimodal scope.}
LFM2-VL and LFM2-Audio are the first multimodal extensions of the LFM2 backbone, aimed at vision and speech on edge devices. Their current capabilities reflect design choices made for on-device deployment.

LFM2-VL targets multi-image vision–language tasks. It supports images up to 512$\times$512 pixels directly and uses tiling plus token reduction for higher resolutions to trade off accuracy against latency and memory. The current model is trained without explicit grounding signals (e.g., bounding boxes, masks, and keypoints) and does not natively handle video or other continuous visual inputs, so we expect weaker performance on fine-grained localization and temporal reasoning. Future work includes adding lightweight grounding compatible with edge budgets, expanding the supported resolution range, exploring video extensions, and integrating sensor streams. We also plan post-training stages for the VLM with reinforcement-learning-style objectives to improve visual reasoning and long-horizon instruction following.

LFM2-Audio focuses on speech applications such as ASR, TTS, and conversational assistants, using a mixture of transcription, classification, TTS, and audio-chat data. We do not systematically evaluate non-speech audio (e.g., music or environmental sounds) or overlapping multi-speaker speech, and training is dominated by English and other high-resource languages, which likely reduces robustness for low-resource languages and accents. Future work includes expanding to richer multilingual and non-speech audio corpora, studying streaming behavior under real device constraints, and exploring on-device personalization.

\paragraph{Late Interaction Models}
While European languages show strong mutual transfer, Arabic and East Asian languages (JA, KO, ZH) exhibit weaker cross-lingual performance. Future work should investigate targeted training strategies for distant language pairs, such as multilingual contrastive objectives or language-adversarial training. The current model is trained on general-domain retrieval, however, specialized domains (e.g., biomedical literature, legal documents, scientific papers) are likely to benefit from domain-specific adaptation. Thanks to the modular architecture, this can be achieved by continuing to train the projection layer while keeping the LFM2 backbone frozen, reducing the cost of domain-specific deployments. Finally, the current configuration limits documents to 512 tokens, which is sufficient for most retrieval scenarios but restrictive for long-form content. Since the LFM2 backbone supports 32K tokens, extending document encoding to longer sequences could improve retrieval for technical documentation, research papers, and legal contracts, but would require careful management of MaxSim computation and index storage as sequence length grows.

\subsection{Closing Remarks}
LFM2 targets a specific regime: models small enough to run entirely on commodity CPUs and mobile SoCs while still supporting 32K contexts, multimodal inputs, and retrieval. By co-designing architecture, training, and post-training with hardware-in-the-loop, LFM2 models provide strong capabilities for their size, achieve substantially faster prefill and decode on CPUs, while fitting within the RAM budgets of contemporary phones, tablets, and embedded systems. This makes it practical to run assistants, RAG workflows, and speech interfaces fully on-device, without relying on continuous network connectivity or large accelerator clusters.

Shifting more inference from centralized data centers to edge devices has direct implications for privacy, reliability, and resource use. Processing text, images, and audio locally reduces the need to transmit raw user data to remote servers, allowing applications to function in low-connectivity settings. At the same time, reducing reliance on remote inference can substantially lower the aggregate cost of serving everyday interactions. We expect several of the techniques introduced here, from hardware-in-the-loop optimized architectures to improved training methods, lightweight multimodal extensions, and retrieval-focused variants, to be broadly applicable beyond this model family. 

Together with the open weights and deployment recipes (Section \ref{subsed:avail}), we hope LFM2 serves as a strong and practical baseline for building and studying edge-first foundation models.

\subsection{Availability}
\label{subsed:avail}
All models, including the task-specific Liquid-Nanos family, are released with open weights at \href{https://huggingface.co/LiquidAI}{HuggingFace.co/LiquidAI} with optimized deployment packages for ExecuTorch, llama.cpp, and vLLM. The models support deployment on Qualcomm Snapdragon SoCs, AMD Ryzen processors, and standard CPUs. Comprehensive deployment guides and quantization profiles are included to facilitate immediate adoption across diverse hardware platforms.

\section{Authors}
\label{sec:contributors}
\textbf{Contributors (alphabetical by last name):}
Alexander Amini, Anna Banaszak, Harold Benoit, Arthur B\"o\"ok, Tarek Dakhran, Song Duong, Alfred Eng, Fernando Fernandes, Marc H\"ark\"onen, Anne Harrington, Ramin Hasani, Saniya Karwa, Yuri Khrustalev, Maxime Labonne, Mathias Lechner, Valentine Lechner, Simon Lee, Zetian Li, Noel Loo, Jacob Marks, Edoardo Mosca, Samuel J. Paech, Paul Pak, Rom N. Parnichkun, Alex Quach, Ryan Rogers, Daniela Rus, Nayan Saxena, Bettina Schlager, Tim Seyde, Jimmy T.H. Smith, Aditya Tadimeti, Neehal Tumma

\bibliographystyle{plainnat}
\bibliography{references}

\newpage
\appendix


\section{Decoupled Top-K Knowledge Distillation}
\label{app:dtk}

\subsection{Forward KL}

\paragraph{Notation.}
Let \(\mathcal A\) be the vocabulary and \(\mathcal T(x_c)\subset\mathcal A\) the teacher’s Top-K set for context \(x_c\). Write \(\bar{\mathcal T}(x_c)=\mathcal A\setminus \mathcal T(x_c)\). Define
\[
P_T(\mathcal T\mid x_c)=\sum_{x\in\mathcal T(x_c)} P_T(x\mid x_c),\quad
P_S(\mathcal T\mid x_c)=\sum_{x\in\mathcal T(x_c)} P_S(x\mid x_c),
\]
\[
P_T(x\mid \mathcal T,x_c)=\frac{P_T(x\mid x_c)}{P_T(\mathcal T\mid x_c)},\qquad
P_S(x\mid \mathcal T,x_c)=\frac{P_S(x\mid x_c)}{P_S(\mathcal T\mid x_c)}\quad (x\in\mathcal T).
\]
Let \(\mathrm{Bern}(p)\) denote a Bernoulli distribution with success probability \(p\).

\paragraph{Chain-rule identity for forward KL.}
Starting from
\begin{equation}
D_{\mathrm{KL}}\!\big(P_T(\cdot\mid x_c)\,\big\|\,P_S(\cdot\mid x_c)\big)
=\sum_{x\in\mathcal A} P_T(x\mid x_c)\log\frac{P_T(x\mid x_c)}{P_S(x\mid x_c)} ,
\label{eq:kl_start}
\end{equation}
split the sum over \(\mathcal T\) and \(\bar{\mathcal T}\), write \(P_T(x\mid x_c)=P_T(\mathcal T\mid x_c)\,P_T(x\mid\mathcal T,x_c)\) for \(x\in\mathcal T\) (and analogously for \(\bar{\mathcal T}\)), and add--subtract the corresponding membership terms. This yields
\begin{align}
D_{\mathrm{KL}}\!\big(P_T(\cdot\mid x_c)\,\big\|\,P_S(\cdot\mid x_c)\big)
&= D_{\mathrm{KL}}\!\big(\mathrm{Bern}(P_T(\mathcal T\mid x_c))\,\big\|\,\mathrm{Bern}(P_S(\mathcal T\mid x_c))\big) \nonumber\\
&\ \ \ + P_T(\mathcal T\mid x_c)\,D_{\mathrm{KL}}\!\big(P_T(\cdot\mid \mathcal T,x_c)\,\big\|\,P_S(\cdot\mid \mathcal T,x_c)\big) \nonumber\\
&\ \ \ + P_T(\bar{\mathcal T}\mid x_c)\,D_{\mathrm{KL}}\!\big(P_T(\cdot\mid \bar{\mathcal T},x_c)\,\big\|\,P_S(\cdot\mid \bar{\mathcal T},x_c)\big).
\label{eq:kl_chain_rule}
\end{align}

\paragraph{Decoupled Top-K objective (lower bound).}
Because teacher logits on \(\bar{\mathcal T}(x_c)\) are unavailable, we drop the last (non-negative) term and optimize the lower bound
\begin{equation}
\mathcal L_{\text{DTK}}^{LB}(x_c)
= D_{\mathrm{KL}}\!\big(\mathrm{Bern}(P_T(\mathcal T\mid x_c))\,\big\|\,\mathrm{Bern}(P_S(\mathcal T\mid x_c))\big)
+ P_T(\mathcal T\mid x_c)\,D_{\mathrm{KL}}\!\big(P_T(\cdot\mid \mathcal T,x_c)\,\big\|\,P_S(\cdot\mid \mathcal T,x_c)\big).
\label{eq:dtk_lower_bound}
\end{equation}

\paragraph{Temperature placement.}
We apply temperature only to the conditional Top-K distributions $P_T(\cdot\mid \mathcal T,x_c)$ and $P_S(\cdot\mid \mathcal T,x_c)$, i.e., we replace the conditional KL in Eq.~\eqref{eq:dtk_lower_bound} by $D_{\mathrm{KL}}^{(\tau)}$, while keeping both the Bernoulli term and the scalar weight $P_T(\mathcal T\mid x_c)$ untempered. For \(\tau\ge 1\),
\begin{equation}
p^{(\tau)}(x)=\frac{p(x)^{1/\tau}}{\sum_{y} p(y)^{1/\tau}},\qquad
D_{\mathrm{KL}}^{(\tau)}(p\|q)=\tau^2\,D_{\mathrm{KL}}\!\big(p^{(\tau)}\|q^{(\tau)}\big),
\label{eq:temperature_def}
\end{equation}
giving
\begin{equation}
\boxed{\;
\mathcal L_{\text{DTK}}(x_c)
= D_{\mathrm{KL}}\!\big(\mathrm{Bern}(P_T(\mathcal T\mid x_c))\,\big\|\,\mathrm{Bern}(P_S(\mathcal T\mid x_c))\big)
+ P_T(\mathcal T\mid x_c)\,D_{\mathrm{KL}}^{(\tau)}\!\big(P_T(\cdot\mid \mathcal T,x_c)\,\big\|\,P_S(\cdot\mid \mathcal T,x_c)\big)
\;}
\label{eq:dtk_final}
\end{equation}
with the membership term untempered.

\paragraph{Interpretation.}
The first term (\(\mathcal L_B\)) matches the \emph{amount} of mass on the Top-K, while the second term (\(\mathcal L_T\)) matches the \emph{shape} within the Top-K.

\paragraph{Why naive Top-K+temperature is unstable.}
Let $\hat P_T$ denote the teacher distribution truncated to $\mathcal T(x_c)$ (zero mass outside $\mathcal T$). If we first truncate the teacher to $\hat P_T$ and then apply temperature over the full vocabulary $\mathcal A$, we obtain, as $\tau\to\infty$,
\begin{equation}
\lim_{\tau\to\infty} P_S^{(\tau)}(x)=\tfrac{1}{|\mathcal A|}\quad(\forall x\in\mathcal A),\qquad
\lim_{\tau\to\infty} \hat P_T^{(\tau)}(x)=
\begin{cases}
\tfrac{1}{|\mathcal T|}, & x\in\mathcal T,\\[2pt]
0, & x\notin\mathcal T,
\end{cases}
\label{eq:naive_instability}
\end{equation}
so $D_{\mathrm{KL}}(\hat P_T^{(\tau)}\|P_S^{(\tau)})\to \log \frac{|\mathcal A|}{|\mathcal T|}$ and, by definition $D_{\mathrm{KL}}^{(\tau)}=\tau^2 D_{\mathrm{KL}}(\hat P_T^{(\tau)}\|P_S^{(\tau)})$, the tempered KL grows as $\tau^2$. In contrast, with full logits both tempered distributions converge to uniform on $\mathcal A$ and the KL tends to zero. Conditioning both teacher and student on $\mathcal T$ and tempering only these conditional distributions makes them both converge to uniform on $\mathcal T$, removing this support mismatch.

\subsection{Reverse KL variant}
\label{app:dtk_reverse}
For completeness, we also include some thoughts on the derivation of a decoupled reverse KL.
\paragraph{Reverse KL chain rule.}
With \(Z=\mathbf 1[x\in\mathcal T(x_c)]\), the reverse KL decomposes as
\begin{align}
D_{\mathrm{KL}}\!\big(P_S(\cdot\mid x_c)\,\big\|\,P_T(\cdot\mid x_c)\big)
&= D_{\mathrm{KL}}\!\big(\mathrm{Bern}(P_S(\mathcal T\mid x_c))\,\big\|\,\mathrm{Bern}(P_T(\mathcal T\mid x_c))\big) \nonumber\\
&\quad + P_S(\mathcal T\mid x_c)\,D_{\mathrm{KL}}\!\big(P_S(\cdot\mid \mathcal T,x_c)\,\big\|\,P_T(\cdot\mid \mathcal T,x_c)\big) \nonumber\\
&\quad + P_S(\bar{\mathcal T}\mid x_c)\,D_{\mathrm{KL}}\!\big(P_S(\cdot\mid \bar{\mathcal T},x_c)\,\big\|\,P_T(\cdot\mid \bar{\mathcal T},x_c)\big).
\label{eq:rev_chain}
\end{align}

\paragraph{Lower bound with Top-K.}
Because teacher logits on \(\bar{\mathcal T}(x_c)\) are unavailable, we drop the last (non-negative) term and optimize the computable lower bound
\begin{equation}
\mathcal L_{\text{rev}}^{\text{(LB)}}(x_c)
= D_{\mathrm{KL}}\!\big(\mathrm{Bern}(P_S(\mathcal T\mid x_c))\,\big\|\,\mathrm{Bern}(P_T(\mathcal T\mid x_c))\big)
+ P_S(\mathcal T\mid x_c)\,D_{\mathrm{KL}}^{(\tau)}\!\big(P_S(\cdot\mid \mathcal T,x_c)\,\big\|\,P_T(\cdot\mid \mathcal T,x_c)\big).
\label{eq:rev_lb}
\end{equation}
As in the forward-KL case, we apply temperature \(\tau\) only to the conditional Top-K term:
\[
p^{(\tau)}(x)=\frac{p(x)^{1/\tau}}{\sum_{y} p(y)^{1/\tau}},\qquad
D_{\mathrm{KL}}^{(\tau)}(p\|q)=\tau^2\,D_{\mathrm{KL}}\!\big(p^{(\tau)}\|q^{(\tau)}\big).
\]

\paragraph{Teacher-weighted surrogate.}
To avoid incentivizing the student to down-weight \(P_S(\mathcal T\mid x_c)\), we also consider a teacher-weighted surrogate:
\begin{equation}
\boxed{\;
\mathcal L_{\text{rev}}^{\text{(tw)}}(x_c)
= D_{\mathrm{KL}}\!\big(\mathrm{Bern}(P_S(\mathcal T\mid x_c))\,\big\|\,\mathrm{Bern}(P_T(\mathcal T\mid x_c))\big)
+ P_T(\mathcal T\mid x_c)\,D_{\mathrm{KL}}^{(\tau)}\!\big(P_S(\cdot\mid \mathcal T,x_c)\,\big\|\,P_T(\cdot\mid \mathcal T,x_c)\big)
\;}
\label{eq:rev_tw}
\end{equation}
This variant is no longer a strict lower bound, but empirically removes the incentive to shrink \(P_S(\mathcal T\mid x_c)\). As before, the outer binary term is untempered and temperature is applied only to the Top-K conditional term.

\section{Model Merging Techniques}
\label{app:model_merging}

All merging methods operate on model state dictionaries loaded from safetensors format and merge parameters tensor-by-tensor while preserving the original dtype of each parameter.

\paragraph{Model Soup}
The simplest approach performs weighted averaging of corresponding parameters across models~\citep{wortsman2022modelsoupsaveragingweights}. Given $n$ models with parameters $\theta_1, \ldots, \theta_n$ and normalized weights $w_1, \ldots, w_n$ (where $\sum_{i=1}^n w_i = 1$), the merged parameters are computed as:
\begin{equation}
\theta_{\text{merged}} = \sum_{i=1}^n w_i \theta_i
\end{equation}

\paragraph{Task Arithmetic}
Task arithmetic~\citep{ilharco2023editingmodelstaskarithmetic} extends model soups by working with parameter deltas relative to a base model. For a base model with parameters $\theta_0$ and fine-tuned models $\theta_1, \ldots, \theta_n$, we first compute the task vectors $\tau_i = \theta_i - \theta_0$ representing the adaptations learned during fine-tuning. The merged model is then constructed as:
\begin{equation}
\theta_{\text{merged}} = \theta_0 + \sum_{i=1}^n w_i \tau_i
\end{equation}
This formulation allows us to control the contribution of each fine-tuned capability independently.

\paragraph{TIES-Merging}
TIES (Trim, Elect Sign \& Merge)~\citep{yadav2023tiesmergingresolvinginterferencemerging} addresses parameter interference in task arithmetic through a three-step procedure. First, magnitude-based sparsification retains only the top $k\%$ of parameters by absolute value in each task vector $\tau_i$, setting the remainder to zero. Second, sign election resolves conflicts by computing the majority sign for each parameter position across all task vectors. Third, the disjoint merge averages only those parameters that agree with the elected majority sign. Formally, given sparsified task vectors $\tilde{\tau}_i$ and elected signs $\gamma_j = \text{sign}\left(\sum_{i=1}^n \tilde{\tau}_{ij}\right)$, we construct a consensus mask $M_{ij} = \mathbb{1}[\text{sign}(\tilde{\tau}_{ij}) = \gamma_j]$ and compute:
\begin{equation}
\theta_{\text{merged}} = \theta_0 + \sum_{j} \frac{\sum_{i=1}^n w_i \tilde{\tau}_{ij} M_{ij}}{\sum_{i=1}^n w_i M_{ij}}
\end{equation}
where the denominator normalizes by the sum of weights that agree with the majority sign.

\paragraph{DARE}
DARE (Drop And REscale)~\citep{yu2024languagemodelssupermario} employs random sparsification as an alternative to magnitude-based pruning. Each parameter in a task vector is dropped with probability $p$ (the drop rate) and retained otherwise. To maintain the expected magnitude of the merged updates, retained parameters are rescaled by $1/(1-p)$. For a task vector $\tau_i$ and drop rate $p$, the sparsified vector is computed as:
\begin{equation}
m_i \sim \text{Bernoulli}(p), \quad \tilde{\tau}_i = (1 - m_i) \odot \tau_i, \quad \hat{\tau}_i = \frac{\tilde{\tau}_i}{1-p}
\end{equation}
where $\odot$ denotes element-wise multiplication. We implement DARE in combination with both standard task arithmetic (DARE-linear) and with TIES-style consensus (DARE-TIES), where the random sparsification replaces the magnitude-based trimming step while preserving the sign election and disjoint merge procedures.

\paragraph{DELLA}
DELLA (Drop and rEscaLe via sampLing with mAgnitude)~\citep{deep2024dellamergingreducinginterferencemodel} extends DARE by replacing uniform random dropout with magnitude-aware sampling. The key insight is that lower-magnitude parameters contribute less to task-specific performance and can tolerate higher dropout rates. For each layer, parameters are ranked by absolute magnitude, and dropout probabilities are assigned inversely proportional to rank:
\begin{equation}
r_i = \text{rank}(|\tau_{ij}|), \quad p_i = p_{\min} + \frac{\epsilon}{n} \cdot r_i
\end{equation}
where $p_{\min} = p - \epsilon/2$ ensures the average dropout rate equals $p$, and $\epsilon$ controls the spread of probabilities across ranks. Each parameter is then independently dropped and rescaled:
\begin{equation}
m_i \sim \text{Bernoulli}(p_i), \quad \hat{\tau}_i = \frac{(1 - m_i) \odot \tau_i}{1 - p_i}
\end{equation}
The per-parameter rescaling by $1/(1-p_i)$ preserves the expected contribution of each parameter to the output embeddings. We combine DELLA's magnitude-based sampling with TIES-style sign election and disjoint merging, where parameters agreeing with the majority sign are averaged after the stochastic pruning step.

\section{Evaluation Details}
\label{app:evaluation}

The evaluation scores for the text-only models come from a custom internal harness. We designed the benchmark implementations to reliably extract final answers from model outputs, whereas open-source evaluation harnesses use more restrictive parsers that may miss correct answers due to formatting variations.

\subsection*{MMLU}

Based on Hendrycks et al., Measuring Massive Multitask Language Understanding~\citep{hendrycks2021mmlu}.

\begin{itemize}
    \item \textbf{Dataset:} \texttt{cais/mmlu}
    \item \textbf{Implementation:} 5-shot by default (uses first 5 examples from dev set per subject). We apply the chat template to build a multi-turn conversation with each shot as instruction-answer pair.
    \item \textbf{Parsing:} Takes the highest logit token, strips whitespace/newlines, falls back to constraint-based prediction (max probability among A/B/C/D) if the stripped token is not a valid choice. This makes it more robust to tokenizer differences.
    \item \textbf{Scoring:} Constrains to A/B/C/D token logits via softmax for probability distribution.
\end{itemize}

\subsection*{MMLU-Pro}

From Wang et al., MMLU-Pro: A More Robust and Challenging Multi-Task Language Understanding Benchmark~\citep{wang2024mmlupro}.

\begin{itemize}
    \item \textbf{Dataset:} \texttt{TIGER-Lab/MMLU-Pro}
    \item \textbf{Implementation:} 5-shot by default using Chain-of-Thought (CoT) examples from validation set as a conversation with a chat format applied. We remove all ``N/A'' options from the dataset.
    \item \textbf{Prompt:} ``The following are multiple-choice questions (with answers) about {subject}. Think step by step and then finish your answer with "The answer is (X)" where X is the correct letter choice.\textbackslash n'' as a system prompt.
    \item \textbf{Parsing:} Custom parser with priority order:
    \begin{itemize}
        \item Letter patterns at beginning of last 5 lines (e.g., ``A.'', ``A)'', ``A/'', ``A:'')
        \item ``answer is (X)'' patterns
        \item Standalone parenthesized letters
        \item ``choice X'' or ``option X'' patterns
        \item ``final answer: X'' patterns
        \item Fallback to any letter followed by delimiter
        \item Excludes false positives like ``A: Let's think step by step''
    \end{itemize}
    \item \textbf{Generation:} Greedy decoding, max 8192 tokens.
    \item \textbf{Scoring:} String matching of extracted letter vs ground truth.
\end{itemize}

\subsection*{GPQA Diamond}

From Rein et al., GPQA: A Graduate-Level Google-Proof Q\&A Benchmark~\citep{rein2023gpqa}.

\begin{itemize}
    \item \textbf{Dataset:} \texttt{Idavidrein/gpqa} (\texttt{gpqa\_diamond} subset)
    \item \textbf{Implementation:} Each question evaluated 10 times with different answer orderings to reduce position bias. Chat template is applied.
    \item \textbf{Prompt:} ``What is the correct answer to this question: \{question\}\textbackslash n\textbackslash n\{choices\}"
    \item \textbf{Parsing:} Same approach as in MMLU.
    \item \textbf{Scoring:} Final accuracy is average across all permuted evaluations.
\end{itemize}

\subsection*{IFEval}

From Zhou et al., Instruction-Following Evaluation for Large Language Models~\citep{zhou2023ifeval}.

\begin{itemize}
    \item \textbf{Dataset:} \texttt{google/IFEval}
    \item \textbf{Implementation:} Uses official IFEval reference implementation.
    \item \textbf{Generation:} Greedy decoding, \texttt{max\_tokens=4096}
    \item \textbf{Scoring:} Average of all 4 metrics (\texttt{strict\_prompt}, \texttt{loose\_prompt}, \texttt{strict\_instruction}, \texttt{loose\_instruction})
\end{itemize}

\subsection*{IFBench}

From Pyatkin et al., Generalizing Verifiable Instruction Following~\citep{pyatkin2025ifbench}.

\begin{itemize}
    \item \textbf{Dataset:} \texttt{allenai/IFBench\_test}
    \item \textbf{Implementation:} Uses official IFBench reference implementation.
    \item \textbf{Generation:} Greedy decoding, max 4096 tokens.
    \item \textbf{Scoring:} Similar to IFEval - strict/loose modes, prompt-level and instruction-level accuracy
\end{itemize}

\subsection*{Multi-IF}

From He et al., Multi-IF: Benchmarking LLMs on Multi-Turn and Multilingual Instruction Following~\citep{he2024multiif}. 

\begin{itemize}
    \item \textbf{Dataset:} \texttt{facebook/Multi-IF}
    \item \textbf{Implementation:} Uses official Multi-IF reference implementation. Automatically truncates prompts exceeding model's max context length.
    \item \textbf{Generation:} Greedy decoding, 32768 tokens.
    \item \textbf{Scoring:} Average accuracy across all 3 turns.
\end{itemize}

\subsection*{GSM8K}

From Cobbe et al., Training Verifiers to Solve Math Word Problems~\citep{cobbe2021gsm8k}.

\begin{itemize}
    \item \textbf{Dataset:} \texttt{openai/gsm8k}
    \item \textbf{Implementation:} 5-shot examples randomly sampled from train set. No system message but the chat template is applied to the entire conversation.
    \item \textbf{Prompt:}
    \begin{itemize}
        \item Gemma 3 uses an additional instruction to improve performance ``Solve the following math problem and write the final solution in \textbackslash boxed\{\}\textbackslash n\textbackslash n\{problem\}''
        \item Llama 3.2 uses an additional instruction to improve performance ``Given the following problem, reason and give a final answer to the problem.\textbackslash nYour response should end with "The final answer is [answer]" where [answer] is the response to the problem.''
    \end{itemize}
    \item \textbf{Parsing:} Custom parser with priority order:
    \begin{enumerate}
        \item \texttt{\textbackslash boxed\{\}} pattern (LaTeX)
        \item \texttt{\#\#\#} or \texttt{\#\#\#\#} patterns
        \item ``ANSWER:'' pattern
        \item ``answer is \{\}'' pattern
        \item Other patterns (final answer, etc.)
        \item Numeric fallback from end of text
    \end{enumerate}
    \item \textbf{Generation:} Greedy decoding, max 4096 tokens.
    \item \textbf{Scoring:} Reports both strict (exact match after extraction) and loose (substring match) accuracy. Loose score is used because of low performance with strict scoring with Gemma 3 models.
\end{itemize}

\subsection*{GSMPlus}

From Li et al., GSM-Plus: A Comprehensive Benchmark for Evaluating the Robustness of LLMs as Mathematical Problem Solvers~\citep{li2024gsmplus}.

\begin{itemize}
    \item \textbf{Dataset:} \texttt{qintongli/GSM-Plus}
    \item \textbf{Implementation:} Same as GSM8K.
    \item \textbf{Prompt:} Same as GSM8K.
    \item \textbf{Parsing:} Same as GSM8K.
    \item \textbf{Generation:} Greedy decoding, max 4096 tokens.
    \item \textbf{Scoring:} Strict matching only.
\end{itemize}

\subsection*{MATH 500}

MATH-500 is a 500-problem subset of MATH constructed in the Let’s Verify Step by Step paper~\citep{lightman2023letsverify}.

\begin{itemize}
    \item \textbf{Dataset:} \texttt{HuggingFaceH4/MATH-500}
    \item \textbf{Implementation:} 4-shot fixed examples with chat template applied to the entire conversation.
    \item \textbf{Prompt:} ``Problem:\textbackslash n\{problem\}''
    \item \textbf{Parsing:} Same as GSM8K.
    \item \textbf{Generation:} Greedy decoding, max 4096 tokens.
\end{itemize}

\subsection*{MATH Lvl 5}

Level 5 problems come from the original MATH dataset~\citep{hendrycks2021math}.

\begin{itemize}
    \item \textbf{Dataset:} \texttt{EleutherAI/hendrycks\_math} (Level 5 only)
    \item \textbf{Implementation:} 0-shot with chat template.
    \item \textbf{Prompt:} ``Problem:\textbackslash n\{problem\}''
    \item \textbf{Parsing:} Same as GSM8K.
    \item \textbf{Generation:} Greedy decoding, max 4096 tokens.

\end{itemize}

\subsection*{MMMLU}

MMMLU is a multilingual extension of MMLU released by OpenAI~\citep{openai2024mmmlu}.

\begin{itemize}
    \item \textbf{Dataset:} \texttt{openai/MMMLU}
    \item \textbf{Implementation:} 0-shot without chat template. Seven language subsets - Arabic (AR\_XY), German (DE\_DE), Spanish (ES\_LA), French (FR\_FR), Japanese (JA\_JP), Korean (KO\_KR), Chinese (ZH\_CN).
    \item \textbf{Parsing:} Same as MMLU.
    \item \textbf{Scoring:} Reports average across all 7 languages.
\end{itemize}

\subsection*{MGSM}

From Shi et al., Language Models are Multilingual Chain-of-Thought Reasoners~\citep{shi2023mgsm}.

\begin{itemize}
    \item \textbf{Dataset:} \texttt{juletxara/mgsm}
    \item \textbf{Implementation:} 5-shot examples from training dataset per language. Six languages - Spanish (es), French (fr), German (de), Chinese (zh), Japanese (ja), Russian (ru).
    \item \textbf{Prompt:} Same as GSM8K.
    \item \textbf{Parsing:} Same as GSM8K.
    \item \textbf{Generation:} Greedy decoding, max 4096 tokens.
    \item \textbf{Scoring:} Reports average loose accuracy across all 6 languages. Both strict and loose accuracy per language.
\end{itemize}

\clearpage
\section{Multilingual Vision Evaluations}
\label{app:vl-multilingual}

We provide detailed per-language multilingual vision evaluation results in Table \ref{tab:vl-multilingual}. All multilingual benchmarks were translated from English into Arabic, Chinese, French, German, Italian, Japanese, Korean, Portuguese, and Spanish using GPT-4.1-mini.

\begin{table}[h!]
\centering
\scriptsize
\begin{tabular}{lccccc}
\toprule
\textbf{Language} & \textbf{SmolVLM2-2.2B} & \textbf{InternVL3\_5-2B} & \textbf{Qwen3-VL-2B} & \textbf{Qwen2.5-VL-3B} & \textbf{LFM2-VL-3B} \\
\midrule
\# Total Params & 2.2B & 2.3B & 2.1B & 3.75B & 3.0B  \\
\midrule
\multicolumn{6}{c}{\textit{Multilingual MMBench}} \\
\midrule
Arabic & 21.56 & 60.78 & 62.33 & 69.87 & 73.16 \\
Chinese & 50.30 & 73.16 & 74.29 & 75.41 & 74.37 \\
French & 53.07 & 71.95 & 72.47 & 75.58 & 77.14 \\
German & 47.62 & 72.29 & 72.47 & 75.50 & 78.18 \\
Italian & 47.01 & 70.65 & 70.65 & 75.15 & 77.14 \\
Japanese & 40.78 & 68.92 & 69.26 & 73.94 & 74.98 \\
Korean & 28.23 & 68.23 & 67.97 & 71.95 & 74.46 \\
Portuguese & 52.29 & 71.86 & 63.64 & 74.72 & 76.36 \\
Spanish & 55.76 & 72.47 & 72.64 & 76.88 & 76.80 \\
\midrule
\textbf{Average} & 44.07 & 70.03 & 69.52 & 74.33 & 75.84 \\
\midrule
\multicolumn{6}{c}{\textit{MMMB}} \\
\midrule
Arabic & 43.69 & 70.10 & 71.01 & 73.74 & 81.31 \\
Chinese & 64.75 & 79.49 & 77.73 & 79.90 & 80.40 \\
French & 64.49 & 77.32 & 73.13 & 78.74 & 82.63 \\
German & 59.44 & 75.66 & 75.66 & 78.84 & 82.63 \\
Italian & 61.06 & 76.57 & 73.69 & 78.43 & 82.22 \\
Japanese & 56.67 & 75.40 & 74.14 & 76.26 & 80.71 \\
Korean & 46.87 & 73.03 & 72.47 & 75.45 & 80.86 \\
Portuguese & 59.80 & 77.02 & 72.58 & 77.98 & 81.06 \\
Spanish & 63.64 & 78.18 & 76.57 & 79.04 & 81.87 \\
\midrule
\textbf{Average} & 57.82 & 75.86 & 74.11 & 77.60 & 81.52 \\
\bottomrule
\end{tabular}
\caption{Performance of 2--4B VLMs on multilingual MMBench and MMMB. All results are obtained using VLMEvalKit~\citep{duan2025vlmevalkitopensourcetoolkitevaluating}.}
\label{tab:vl-multilingual}
\end{table}

\end{document}